%% file: 0038.tex
\documentclass[runningheads]{llncs}
\usepackage{graphicx}
\usepackage{amsmath,amssymb} 
\usepackage{color}
\usepackage{subfigure}
\usepackage[colorlinks,linkcolor=red]{hyperref}

\newcommand{\etal}{\textit{et al.}}

\makeatletter
\newcommand{\printfnsymbol}[1]{%
  \textsuperscript{\@fnsymbol{#1}}%
}
\makeatother

\makeatletter
\renewcommand{\paragraph}{%
  \@startsection{paragraph}{4}%
  {\z@}{1ex \@plus 1ex \@minus .2ex}{-1em}%
  {\normalfont\normalsize\bfseries}%
}
\makeatother

\begin{document}
\pagestyle{headings}
\mainmatter

\def\ACCV20SubNumber{38}

\title{Class-incremental Learning with \\Rectified Feature-Graph Preservation} 
\titlerunning{Class-incremental Learning with \\Rectified Feature-Graph Preservation}

\author{Cheng-Hsun Lei\thanks{Both authors contributed equally to the paper} \and
Yi-Hsin Chen\printfnsymbol{1} \and
Wen-Hsiao Peng \and 
Wei-Chen Chiu
}

\authorrunning{Lei and Chen et al.}

\institute{National Chiao Tung University, Taiwan\\
\email{\{raygoah.cs07g, yhchen.iie07g\}@nctu.edu.tw} 
\email{\{wpeng, walon\}@cs.nctu.edu.tw}
}

\maketitle
\begin{abstract}
In this paper, we address the problem of distillation-based class-incremental learning with a single head. A central theme of this task is to learn new classes that arrive in sequential phases over time while keeping the model's capability of recognizing seen classes with only limited memory for preserving seen data samples. Many regularization strategies have been proposed to mitigate the phenomenon of catastrophic forgetting. To understand better the essence of these regularizations, we introduce a feature-graph preservation perspective. Insights into their merits and faults motivate our weighted-Euclidean regularization for old knowledge preservation. We further propose rectified cosine normalization and show how it can work with binary cross-entropy to increase class separation for effective learning of new classes. Experimental results on both CIFAR-100 and ImageNet datasets demonstrate that our method outperforms the state-of-the-art approaches in reducing classification error, easing catastrophic forgetting, and encouraging evenly balanced accuracy over different classes. Our project page is at : \href{https://github.com/yhchen12101/FGP-ICL}{https://github.com/yhchen12101/FGP-ICL}. 
\end{abstract}

\input{introduction.tex}
\input{related.tex}

\input{method.tex}
\input{experiment.tex}
\input{conclusion.tex}

\bibliographystyle{splncs}
\bibliography{egbib}
\clearpage

\appendix
\title{Appendix : Class-incremental Learning\\ with Rectified Feature-Graph Preservation\vspace{-1em}} 
\author{}
\institute{}
\authorrunning{Lei and Chen et al.}
\titlerunning{Class-incremental Learning with Rectified Feature-Graph Preservation}
\maketitle
\input{appendix.tex}

\end{document}

%% file: introduction.tex
\section{Introduction}
\label{sec:introduction}
    
Class-incremental learning~\cite{rebuffi2017icarl} is a difficult yet practical problem for visual recognition models. The task requires the model with a single classification head to learn new classes that arrive sequentially while preserving its knowledge on the old ones.
It is worth noting that class-incremental learning is different from task-incremental learning~\cite{aljundi2018memory,kirkpatrick2017overcoming,li2017learning,hou2018lifelong,aljundi2017expert}, which learns to handle multiple tasks (i.e. groups of classes, possibly from different datasets) given sequentially. In particular, task-incremental learning has a unique assumption that the task indices of classes are known during both training and test time.  Thus, the model typically aims to learn a shared feature extractor across tasks under the multi-head setup (i.e. each task has its own head). Even though there are some works~\cite{kirkpatrick2017overcoming,lopez2017gradient,chaudhry2018efficient} addressing class-incremental learning from the perspective of task-incremental learning, the differences stemmed from the aforementioned assumption clearly distinguish these works from typical class-incremental learning. This paper focuses on the more difficult yet practical setup to learn a single-head classifier without task identification.

Generally, class-incremental learning is faced with two major challenges: (1) the model needs to be updated on-the-fly upon the arrival of training data of new classes, and (2) there is only limited memory space (also known as experience reply buffer) for retaining partially the training data of previously learned classes. It is shown in \cite{rebuffi2017icarl} that naively fine-tuning the model on newly received training data of incoming classes without considering the previously learned classes often suffers from \emph{catastrophic forgetting} \cite{mccloskey1989catastrophic}; that is, the classification accuracy on old classes may decline dramatically when the model attempts to learn new classes.
    
Several prior works based on deep learning models have been proposed to deal with catastrophic forgetting. They can roughly be categorized into three types: the parameter-based, distillation-based, and generative-model-based approaches. The parameter-based methods aim to restrict the update on network parameters that are highly sensitive to the recognition accuracy of old classes \cite{aljundi2018memory,kirkpatrick2017overcoming,zenke2017continual,chaudhry2018riemannian}, and to conserve the usage of network parameters in learning new classes~\cite{aljundi2018selfless,mallya2018piggyback,mallya2018packnet}. Identifying and regularizing those important network parameters can be prohibitively expensive, especially with large learning models. By contrast, the distillation-based methods leverage the idea of knowledge distillation~\cite{hinton2015distilling}, requiring that the model updated with the training data of new classes should produce similar predictions to the old model over the previously learned classes  \cite{rebuffi2017icarl,belouadah2018deesil,castro2018end,hou2019learning,jung2018less,li2017learning,wu2019large,liu2020mnemonics}. They typically need a divergence measure to be defined that measures the similarity between the old and new predictions. In particular, this class of methods is often influenced by the amount of data available in the experience reply buffer on which the similarity measurement is conducted. The generative-model-based methods replace the experience replay buffer with deep generative models~\cite{goodfellow2014generative}, so that these generative models can later be utilized to generate synthetic training data for old classes while updating the model to learn new classes ~\cite{shin2017continual,he2018exemplar,ostapenko2019learning}. Obviously, their capacity for learning the distribution of old data crucially affects the quality of generated training data.   

This work presents a new geometry-informed attempt at knowledge distillation for class-incremental learning. Instead of devising new divergences for similarity measurement, our work stresses preserving the geometric structure of the feature space. Specifically, we first (1) analyze the objectives used in the existing distillation-based algorithms from a \textit{feature-graph preservation} perspective. By inspection of their merits and faults, we then (2) propose \textit{weighted-Euclidean regularization} to preserve the knowledge of previously learned classes while providing enough flexibility for the model to learn new classes. Lastly, we (3) introduce \textit{rectified cosine normalization} to learn discriminative feature representations, together with the objective of binary cross-entropy.

%% file: related.tex
\section{Related Work}
\label{sec:related}
\paragraph{\textbf{Parameter-based.}}
Parameter-based methods attempt to identify the network parameters that have a profound impact on the classification accuracy of previously learned classes once altered. These important parameters are encouraged to be fixed during the incremental learning in order to resolve the issue of catastrophic forgetting (e.g. MAS~\cite{aljundi2018memory}, EWC~\cite{kirkpatrick2017overcoming}, SI~\cite{zenke2017continual}, and RWalk~\cite{chaudhry2018riemannian}). Other works such as Rahaf~\etal~\cite{aljundi2018selfless}, Piggy-back~\cite{mallya2018piggyback}, and Packnet~\cite{mallya2018packnet} further advance freeing up the remaining parameters selectively in learning new classes. This is achieved by learning a weighting mask, pruning the network, or adding specifically-designed regularization. Because separate parts of the network can be activated for different tasks, parameter-based methods often find application in task-incremental learning. They however incur heavy computation in identifying and regularizing important network parameters.

\paragraph{\textbf{Generative-model-based.}}
Generative-model-based methods (e.g. \cite{shin2017continual,he2018exemplar,ostapenko2019learning}) turn to generative models as a means of keeping the training data of old classes. Since the experience replay buffer has a limited size, the training data of previously learned classes can only be retrained partially. This leads to an imbalanced distribution of training data between the new and old classes. To overcome this issue, generative-model-based methods employ generative models to generate synthetic training data for old classes in the course of incremental learning. It is worth noting that the use, storage, and training of these generative models can cause a considerable complexity increase.

\paragraph{\textbf{Distillation-based.}}
Knowledge distillation~\cite{hinton2015distilling} is originally proposed to distill the knowledge from a teacher network (which is typically a network with larger capacity) and transfer it to a student network (which is a smaller network than the teacher), by encouraging the student to have a similar posterior distribution to the teacher's. This concept is adapted to preserve the model's knowledge of old classes for class-incremental learning. LwF~\cite{li2017learning} is the first work to apply knowledge distillation to incremental learning, introducing a modified cross-entropy loss function.
\textbf{iCaRL} \cite{rebuffi2017icarl} uses binary cross-entropy as a new distillation loss, provides a mechanism for selecting old-class exemplars, and classifies images with the \emph{nearest-mean-of-exemplars} criterion. 
Castro \etal~\cite{castro2018end} proposes a framework, termed \textbf{End-to-End} in this paper, that adopts the ordinary cross-entropy as the distillation loss in learning end-to-end the feature representation and the classifier. In particular, it introduces balanced fine-tuning as an extra training step to deal with the data imbalance between the old and new classes.
\textcolor{black}{\textbf{BIC} \cite{wu2019large} proposes a bias correction method to address the data imbalance, achieving excellent accuracy on large datasets.} 
\textbf{Hou19} \cite{hou2019learning} \textcolor{black}{proposes} several designs, including the cosine normalization, the less-forget constraint, and the inter-class separation, to encourage the classifier to treat old and new classes more uniformly. \textcolor{black}{Based on \textbf{Hou19}'s training strategies, Liu \etal~\cite{liu2020mnemonics}, termed \textbf{Mnemonics}, further optimize the parameterized exemplars by a bi-level optimization program.} Both \textbf{Hou19} and \textbf{Mnemonics} rely on a well-pretrained feature extractor to obtain superior performance on incremental learning.

%% file: method.tex
\section{Proposed Method}
\label{sec:method}
    \begin{figure}[t]
    \centering
    \includegraphics[width=.9\textwidth]{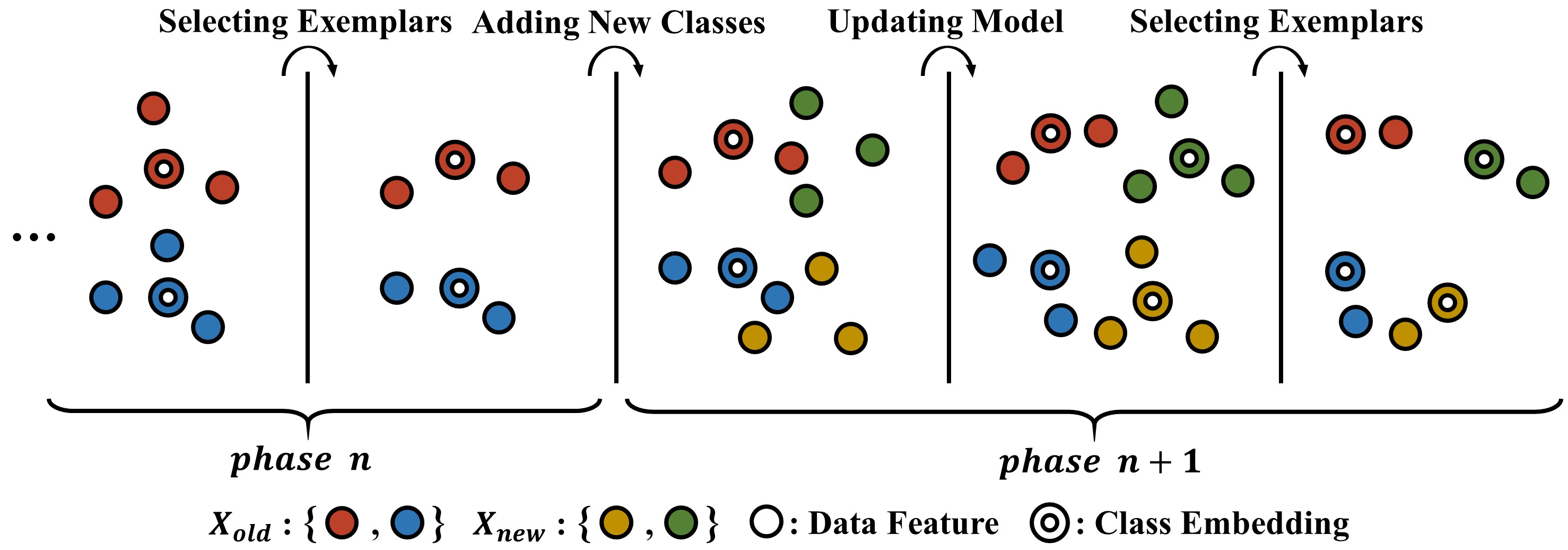}
     \caption{Illustration of the class-incremental learning process: (1) selecting exemplars, (2) adding the data samples of new classes, and (3) updating the model.}
    
    \label{fig:overview}
    \end{figure}
    
    \begin{figure}[t]
    \centering
    \includegraphics[width=.8\textwidth]{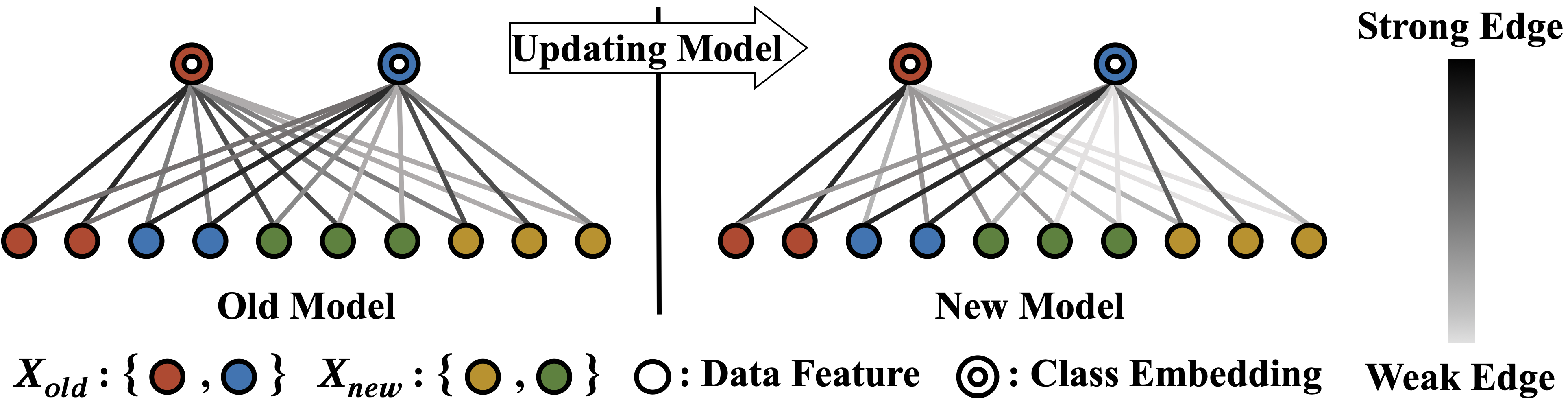}
     \caption{Knowledge distillation from a bipartite feature-graph preservation perspective, with the bottom row and top row denoting the feature vectors of data samples and the class embeddings respectively. Minimizing the distillation loss is viewed as the structure (i.e. edge strength) preservation of such a bipartite graph during the model update.
     }
    
    \label{fig:graph}
    \end{figure}

This section presents our strategies~--~namely, weighted-Euclidean regularization and rectified cosine normalization~--~for dealing with catastrophic forgetting and effective new class learning in the scenario of class-incremental learning. 

Fig.~\ref{fig:overview} illustrates the learning of data features and class embeddings (i.e.~the classifier's weight vectors) from one incremental learning phase $n$ to the next $n+1$. As shown, the data of different classes arrive sequentially in different \textit{phases}. We denote the old data that have arrived before phase $n+1$ as $X_{old}$ and the new data arriving in phase $n+1$ as $X_{new}$. Assuming we already have a model capable of classifying the classes $C_{old}$ in the $X_{old}$, we now aim to train the model to additionally learn the new classes $C_{new}$ in $X_{new}$. However, there is limited memory for storing data samples; we can only preserve some exemplars chosen from the seen samples before phase $n+1$. Thus, the training is based on the dataset $X_{all} = X_{new} \cup X_{exem}$, where $X_{exem}$ is composed of the exemplars of $C_{old}$ (i.e $X_{exem}$ is a tiny subset of $X_{old}$) and the number of exemplars that can be kept is constrained by the memory size. 

\subsection{Mitigating Catastrophic Forgetting}
\label{sec:forgetting}
To motivate the need for our weighted-Euclidean regularization, we introduce a feature-graph preservation perspective on knowledge distillation. This new perspective reveals the essence of existing distillation-based regularization schemes in how they tackle catastrophic forgetting and highlights the striking feature of our method.  

\paragraph{\textbf{Knowledge Distillation in Class-Incremental Learning.}}
\label{sec:framework}
We begin by revisiting the prior works from  iCaRL~\cite{rebuffi2017icarl} and End-to-End~\cite{castro2018end}. To prevent class-incremental learning from catastrophic forgetting, both models introduce a distillation loss that requires the model after the gradient update (termed the \textit{new} model) to produce a similar prediction on $C_{old}$ to that made by the model before the update (termed the \textit{old} model), for any data in $X_{all}$. In symbols, for a data sample $x_k \in X_{all}$, iCaRL~\cite{rebuffi2017icarl} computes the distillation loss $\mathcal{L}_{dist\_bce}(x_k)$ to be the sum of the binary cross-entropy between the old prediction $p^*_{i|k}$ and the new prediction $p_{i|k}$ over different classes $i \in C_{old}$:
    \begin{equation}
    \label{equ:dbce}
    \begin{aligned}
    \mathcal{L}_{dist\_bce}(x_k) = \sum_{i \in C_{old}} \left\{ 1 \times \left[- p^*_{i \vert k} \log p_{i \vert k} - (1-p^*_{i \vert k}) \log (1-p_{i \vert k}) \right]\right\},
    \end{aligned}
    \end{equation}
where $p_{i|k}=\sigma (a_{i|k})$ is the sigmoid output of the activation $a_{i|k}= w^{T}_{i}f_k+b_i$ evaluated based on the new model, with $w_i,b_i$ representing the weight vector and bias of the $i$-th classifier respectively, and $f_k$ denoting the feature vector of $x_k$; and $p^*_{i|k}=\sigma (a^*_{i|k})$ is computed similarly yet with the activation $a^*_{i|k}=w^{*T}_{i} f^*_k+b^*_i$ evaluated with the old model. In contrast, the Eq. (3) in the  End-to-End~\cite{castro2018end} paper minimizes the KL-divergence\footnote{This amounts to minimizing the cross-entropy $\sum_i \{-p^*_{i|k} \times \log p_{i|k}\}$ between $p^*_{i|k}$ and $p_{i|k}$ over $i \in C_{old}$ since $p^*_{i|k}$ is not affected by the model update.} between $p^*_{i|k}$ and $p_{i|k}$ over $i \in C_{old}$ for preserving knowledge learned in previous phases:

    \begin{equation}
    \label{equ:dce}
    \begin{aligned}
     \mathcal{L}_{dist\_KL}(x_k)=\sum_{i \in C_{old}}  \left\{ p^*_{i|k} \times \left[\log p^*_{i|k} - \log p_{i|k} \right]\right\},
    \end{aligned}
    \end{equation}
where $p_{i|k}=softmax(a_{i|k})$ (respectively $p^*_{i|k}=softmax(a^*_{i|k})$) is the softmax output of the activation $a_{i|k}/T$ (respectively, $a^*_{i|k}/T$) attenuated by a temperature parameter $T$. 

Close examination of Eqs.~\eqref{equ:dbce} and \eqref{equ:dce} above suggests a more general distillation loss of the form 
    \begin{equation}
    \label{equ:form}
    \begin{aligned}
     \mathcal{L}_{dist}(x_k)=\sum_{i \in C_{old}}  \left\{\gamma_{i|k} \times \mathcal{D}(p^*_{i|k},p_{i|k}) \right\},
    \end{aligned}
    \end{equation}
with $\mathcal{D}(p^*_{i|k},p_{i|k})$ measuring the discrepancy between the old and new predictions (i.e. $p^*_{i|k}$ and $p_{i|k}$) for class $i$, and $\gamma_{i|k}$ weighting the contribution of $\mathcal{D}(p^*_{i|k},p_{i|k})$ to the resulting loss $\mathcal{L}_{dist}(x_k)$.
\textcolor{black}{Note that a non-zero discrepancy value $\mathcal{D}(p^*_{i|k},p_{i|k})$, be it positive or negative, signals a change in the model's prediction.}
By Eq.~\eqref{equ:form}, iCaRL~\cite{rebuffi2017icarl} is seen to have $\gamma_{i|k}=1$ and $\mathcal{D}(p^*_{i|k},p_{i|k})$ the same as the binary cross-entropy between $p^*_{i|k}$ and $p_{i|k}$, while End-to-End~\cite{castro2018end} sets $\gamma_{i|k}$ and $\mathcal{D}(p^*_{i|k},p_{i|k})$ to $p^*_{i|k}$ and $(\log p^*_{i|k} - \log p_{i|k})$, respectively. 

\paragraph{\textbf{Knowledge Distillation as Feature-Graph Preservation.}} 
A graph interpretation can help us understand better what we are observing from Eq.~\eqref{equ:form}. We first note that Eq.~\eqref{equ:form} is to be \textcolor{black}{optimized towards zero} for all $x_k$ in $X_{all}$ and is concerned with the change in relations between the feature vector $f_k$ of certain $x_k \in X_{all}$ and the class embeddings $w_i,\forall i \in C_{old}$. Their relations are seen to be quantified by $p_{i|k}$; specifically, the higher the value of $p_{i|k}$, the more strongly related the $f_k$ and $w_i$ is. This relationship, as depicted in Fig.~\ref{fig:graph}, can be visualized as a bipartite graph, where the feature vectors $f_k$ of all $x_k \in X_{all}$ form a set of vertices that is disjoint from another set of vertices comprising $w_i,i \in C_{old}$. The graph is bipartite because Eq.~\eqref{equ:form} does not concern the relations among vertices in each of these sets, and each $f_k$ is connected to all the $w_i$ with edge weights specified by $p_{i|k}$. We then \textcolor{black}{view the optimization of the distillation loss over $x_k \in X_{all}$ towards zero} as the structure preservation of such a bipartite graph when the learning progresses from one phase to another. This perspective is motivated by the fact that $\mathcal{D}(p^*_{i|k},p_{i|k})$ in Eq.~\eqref{equ:form} reflects how the strength of an edge changes as a result of the model update in a learning phase. In particular, edge-adaptive preservation is achieved when $\gamma_{i|k}$ is assigned non-uniformly. 

To shed light on the rationale of our design, we now delve deeper into iCaRL~\cite{rebuffi2017icarl} and End-to-End~\cite{castro2018end} in terms of their edge weighting scheme $\gamma_{i|k}$, edge strength specification $p_{i|k}$, and edge discrepancy measure $\mathcal{D}(p^*_{i|k},p_{i|k})$. First, iCaRL~\cite{rebuffi2017icarl} applies equal weighting to all edges on the bipartite graph by having $\gamma_{i|k}=1,\forall i,k$, suggesting that the strength of every edge should be preserved equally after the model update. In contrast, End-to-End~\cite{castro2018end} attaches more importance to stronger edges from the viewpoint of the old model by having $\gamma_{i|k}=p^*_{i|k}$. Second, using sigmoid activation in Eq.~\eqref{equ:dbce}, iCaRL~\cite{rebuffi2017icarl} evaluates the edge strength $p_{i|k}$ as a function of the similarity between $f_k$ and $w_i$ alone, whereas the softmax activation (cp. Eq.~\eqref{equ:dce}) in End-to-End~\cite{castro2018end} renders $p_{i|k}$ dependent on the relative similarities between $f_k$ and all the $w_i,i \in C_{old}$. As such, End-to-End~\cite{castro2018end} imposes the additional constraint $\sum_i \gamma_{i|k}=\sum_i p^*_{i|k}=1$ on edge weighting, implying a dependent weighting scheme across edges connecting $f_k$ and $w_i,i \in C_{old}$. Third, the discrepancy measure $\mathcal{D}(p^*_{i|k},p_{i|k})$ involves $p^*_{i|k}$ and $p_{i|k}$, which in turn depends on the activations $a^*_{i|k}$ and $a_{i|k}$, respectively. These activations reflect explicitly the similarity, or the geometrical relation, between $f_k$ and $w_i$ in the feature space. When viewed as a function of $a_{i|k}$, the $\mathcal{D}(p^*_{i|k},p_{i|k})$ in both iCaRL~\cite{rebuffi2017icarl} and End-to-End~\cite{castro2018end} is non-symmetrical with respect to $a^*_{i|k}$; that is, the same deviation of $a_{i|k}$ from $a^*_{i|k}$ yet in different signs causes $\mathcal{D}(p^*_{i|k},p_{i|k})$ to vary differently.
This implies that their distillation losses may favor some non-symmetrical geometry variations in the feature space.

\paragraph{\textbf{Weighted-Euclidean Regularization.}} 
\label{sec:euclidean}
\textcolor{black}{In designing the distillation loss, we deviate from the divergence-based criterion to adopt a feature-graph preservation approach. This is motivated by our use of the nearest-mean-of-exemplars classifier, which stresses the geometric structure of the feature space.} From the above observations, we conjecture that good regularization for knowledge distillation \textcolor{black}{in the feature space} should possess three desirable properties, including (1) prioritized edge preservation, (2) independent edge strength specification, and (3) $a^*_{i|k}$-symmetric discrepancy measure. The first allows more flexibility in adapting feature learning to new classes while preserving old knowledge, by prioritizing stronger edges. The second dispenses with the constraint that $\sum_i \gamma_{i|k}=\sum_i p^*_{i|k}=1$\textcolor{black}{, since the sum-to-one constraint is essential to a cross-entropy interpretation in End-to-End~\cite{castro2018end}, but is not necessary when $p_{i|k}^*$ is regarded as an edge weighting $\gamma_{i|k}$ from our feature-graph viewpoint}. The third is to penalize equally the positive and negative deviation from $a^*_{i|k}$ for a geometrically-symmetric preservation of the graph in the feature space. To this end, our \textit{weighted-Euclidean regularization} chooses $a_{i|k}$, $p_{i|k}$, $\gamma_{i|k}$ and $D(p^*_{i|k},p_{i|k})$ to be 
    \begin{equation}
    \label{equ:euclidean_factor}
    \begin{aligned}
        a_{i|k} = - \frac{\|\bar{w_i}-\bar{f_k}\|_2^2}{2}&, \qquad
        p_{i|k} = \frac{1}{Z} \exp\{a_{i|k}\},  \qquad
        \gamma_{i|k} = p^*_{i|k} \\ 
        D(p^*_{i|k},p_{i|k}) &= \left(\log p^*_{i|k} - \log p_{i|k} \right)^2 = \left( a^*_{i|k} - a_{i|k} \right)^2    
    \end{aligned}
    \end{equation}
where $\bar{w_i},\bar{f_k}$ are the class embedding and feature vector with rectified cosine normalization (in Sec.~\ref{sec:consine}), $Z$ is a normalization constant, and $a^*_{i|k},p^*_{i|k}$ are the activation and edge strength evaluated with the old model. In particular, we choose the  activation $a_{i|k}= - \|\bar{w_i}-\bar{f_k}\|_2^2/2$ to match the criterion of the nearest-mean-of-exemplars classification \cite{rebuffi2017icarl} at test time. It then follows from Eq.~\eqref{equ:form} that our distillation loss has a form of
    \begin{equation}
    \label{equ:euclidean}
    \mathcal{L}_{dist\_wE}(x_k) =  \sum_{i \in C_{old}} 
    \left\{
    \exp(-\frac{\|\bar{w^*_i}-\bar{f^*_k}\|^2_2}{2}) \times 
    \left( 
    \|\bar{w^*_i}-\bar{f^*_k}\|^2_2 - \|\bar{w_i}-\bar{f_k}\|^2_2 
    \right)^2
    \right\},
    \end{equation}
    where constant terms have been omitted for brevity.
\textcolor{black}{Extensive experiments in Sec.~\ref{sec:ablation_rc_c} validate the deceptively simple design of our distillation loss and confirms its superior performance to the seemingly more principled divergence-based methods (e.g. End-to-End~\cite{castro2018end}).}

\subsection{Learning New Classes}
\label{sec:learn_new}
Learning new classes in the context of incremental learning is often faced with data imbalance. Each new class usually has more training data than an old class since only a limited amount of old data can be stored (as exemplars in the memory). To address this issue, we adopt a two-pronged approach: (1) minimizing binary cross-entropy as the classification loss and (2) applying \emph{rectified cosine normalization} to increase class separation in the feature space.

\paragraph{\textbf{Classification Loss.}}
\label{sec:classification}
In the literature~\cite{castro2018end,hou2019learning,rebuffi2017icarl}, there are two main choices of the classification loss: cross-entropy and binary cross-entropy. The former is mostly used for multi-class classification while the latter is popular for multi-label classification. In the context of class-incremental learning (a kind of multi-class classification), the study in \cite{hou2019learning} shows that the model trained with cross-entropy tends to bias in favor of new classes by increasing their class embeddings $w_i$ and bias terms $b_i$, to take advantage of the imbalanced data distribution for lower classification errors (i.e. the model would quickly adapt to new classes without being able to maintain well the knowledge of old classes). This is because cross-entropy adopts a softmax activation function, which determines the class prediction based on the \textit{relative} strength of the class activations, thereby creating the aforementioned short cut. With the aim of achieving more evenly distributed classification accuracy, we choose the stricter binary cross-entropy to leverage its property that the activation of each class is weighted \textit{absolutely} rather than \textit{relatively} in determining the class prediction. In symbols, it is evaluated as 
    \begin{equation}
    \label{equ:bce}
    \begin{aligned}
    \mathcal{L}_{bce}(x_k) = - \sum_{i \in C_{all}} \Big[\delta_{y_i=y_k}\log(\sigma (a_{i|k}))+\delta_{y_i\neq y_k} \log(1 -\sigma(a_{i|k}))\Big],
    \end{aligned}
    \end{equation}
where $y_k$ is the label of the data sample $x_k$, $C_{all}=C_{old} \cup C_{new}$, and $\delta$ is an indicator function.

    \begin{figure}[t]
    \centering
    \subfigure[$\text{rectified-}CN$]{
    \begin{minipage}[t]{0.23\textwidth}
    \centering
    \includegraphics[width=1\textwidth]{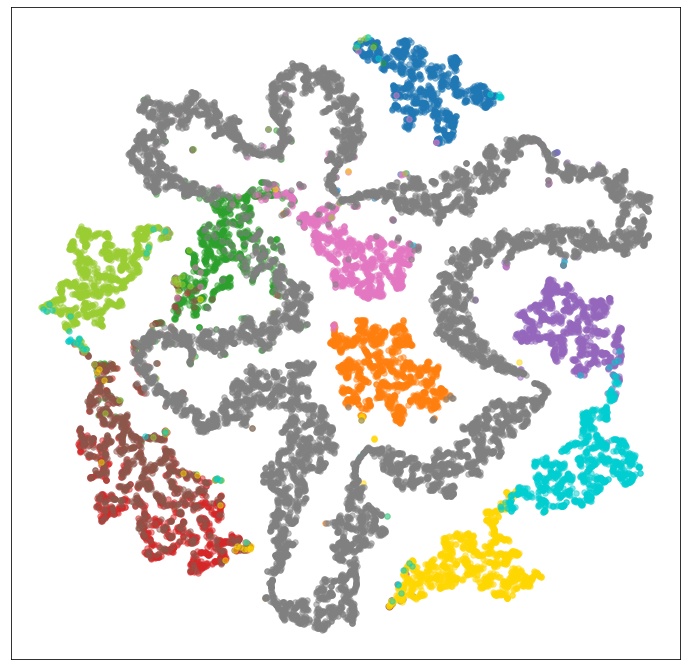}
    \label{fig:with_bias}
    \end{minipage}
    }
    \subfigure[$CN$]{
    \begin{minipage}[t]{0.23\textwidth}
    \centering
    \includegraphics[width=1\textwidth]{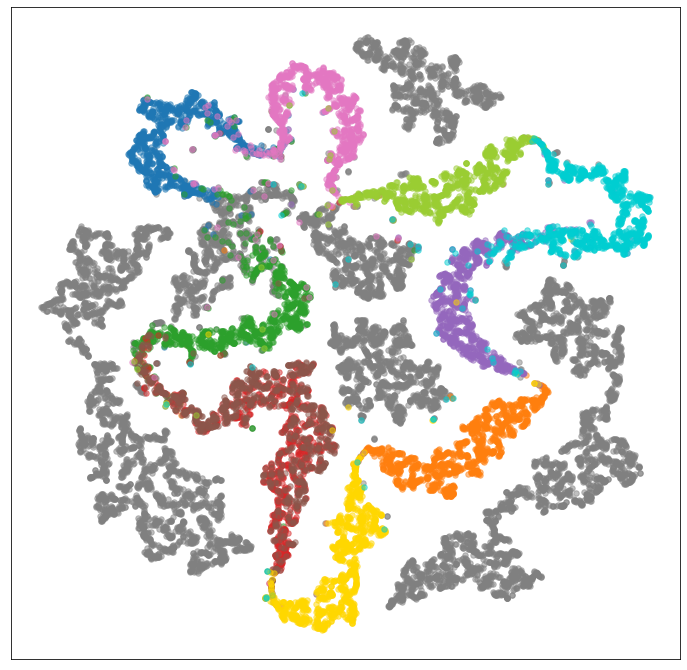} 
    \label{fig:without_bias}
    \end{minipage}
    }
    \caption{
    \textcolor{black}{
    A toy example based on MNIST for showcasing the benefit of our rectified cosine normalization (denoted as $\text{rectified-}CN$) with respect to the typical cosine normalization (denoted as $CN$). The colorized dots in \textbf{(a)} (respectively \textbf{(b)}) are the t-SNE visualization of the feature representations for images of 10 digit classes, which are obtained from the classifier trained with adopting our $\text{rectified-}CN$ (respectively $CN$); while the gray dots are the features from the classifier of using $CN$ (respectively $\text{rectified-}CN$).}}
    \label{fig:MNIST}
    \end{figure}
\label{sec:consine}

\paragraph{\textbf{Rectified Cosine Normalization.}}
In \cite{hou2019learning}, they additionally adopt cosine normalization to address data imbalance by normalizing the class embedding $w_i$ and the feature vector $f_k$ in evaluating the activation ${a}_{i|k} = \bar{w}^{T}_{i}\bar{f_k}$, where $\bar{w}_{i},\bar{f_k}$ are the normalized class embedding and feature vector. \textcolor{black}{Here we further develop a \textit{rectified cosine normalization} technique, which is empirically shown to encourage} greater separation between incrementally-learned classes in the feature space than the typical cosine normalization. 
Our rectified cosine normalization is implemented by (1) augmenting, during training with Eq.~\eqref{equ:bce}, every class embedding $w_i$ with a learnable bias $b_i$ and every feature vector $f_k$ with a constant 1 for separate normalization; and by (2) involving, at test time, only the feature vector $f_k$ (without augmentation) in the nearest-mean-of-exemplars classification \cite{rebuffi2017icarl}.
Step 1 gives rise to an \textit{augmented feature space} of $W_i=(w_i,b_i)$ and $F_k=(f_k,1)$, with the activation\footnote{This activation is to be distinguished from the one $a_{i|k}= - \|\bar{w_i}-\bar{f_k}\|_2^2/2$ for weighted-Euclidean regularization.} evaluated as $a_{i|k}=\bar{W}^{T}_{i}\bar{F_k}$ where the bar indicates normalization.

Here we use MNIST dataset~\cite{lecun1998gradient}, which is composed of images of handwritten digits, to showcase the benefit of our rectified cosine normalization. We construct the classifier by a shallow network (including 2 convolutional layers followed by 3 fully connected layers with an output size of 3, which implies that the feature vector $f$ is three-dimensional), and train it with whole 10 digit classes. Fig.~\ref{fig:MNIST} shows the t-SNE visualization of the feature representations $f$ learned 
from both the model variants of using our rectified cosine normalization and the typical cosine normalization. It is clear to see that our proposed normalization scheme better encourages the separation between classes. In Sec.~\ref{sec:ablation_rc_c}, we present an ablation study to demonstrate the effectiveness of our rectified cosine normalization in improving incremental accuracy.

\subsection{Objective Function}
\label{sec:objective}
\textcolor{black}{
    We combine the binary cross-entropy classification loss, rectified cosine normalization, and the weighted-Euclidean regularization to build our overall objective:
    \begin{equation}
    \label{equ:loss}
    \begin{aligned}
    \mathcal{L}(\mathbf{x}) = \frac{1}{N}\sum^{N}_{k=1}\mathcal({L}_{bce}(x_k)+ \lambda \mathcal{L}_{dist\_wE}(x_k)),
    \end{aligned}
    \end{equation}
    where ${a}_{i|k} = \bar{W}^{T}_{i}\bar{F_k}$ in $\mathcal{L}_{bce}(x_k)$, $\{x_1,x_2,\ldots,x_N\}$ is a mini-batch drawn from $X_{all}$, and $\lambda$ is the hyper-parameter used to balance $\mathcal{L}_{bce}$ against $\mathcal{L}_{dist\_wE}$. Inspired by \cite{Taitelbaum2018AddingNC}, we choose $\lambda$ as follows in recognizing that: in the early stages of incremental learning, when the relative size of old classes to new classes is small, the weighted-Euclidean regularization should be de-emphasized as the feature extractor may not be well learned yet. Note that the ratio $|C_{old}|/|C_{all}|$ is updated constantly along the incremental learning process.
    \begin{equation}
    \label{equ:loss lambda}
    \begin{aligned}
    \lambda = \lambda_{base}\sqrt{\frac{\left|C_{old}\right|}{\left|C_{all}\right|}},
    \end{aligned}
    \end{equation}
    where $\lambda_{base}=0.1$ is set empirically.} 

%% file: experiment.tex
\section{Experimental Results}
\label{sec:experiments}
    \paragraph{\textbf{Datasets and Baselines.}}
    \textcolor{black}{We follow the same experiment protocol as prior works.} Two datasets, CIFAR-100 \cite{krizhevsky2009learning} and ImageNet \cite{deng2009imagenet} with 100 classes selected randomly, are used for experiments. The data samples of these 100 classes comprise a data stream that arrives in a class-incremental manner in 10 phases, each adding 10 new classes to those already received in previous phases. Particularly, we adopt two training scenarios. One is to train the model from scratch, i.e. the model starts with random initialization and learns 100 classes incrementally. The other follows \cite{hou2019learning} to train the model with the first 50 classes in order to have a reasonably good feature extractor to begin with. Then the remaining 50 classes are split evenly and learned in 5 sequential phases. The baselines include iCaRL~\cite{rebuffi2017icarl}, End-to-End~\cite{castro2018end}, \textcolor{black}{BIC~\cite{wu2019large}}, Hou19~\cite{hou2019learning}, and \textcolor{black}{Mnemonics~\cite{liu2020mnemonics}}.
    
    \paragraph{\textbf{Metrics.}}
    The evaluation basically adopts the same common metrics used in prior works, 
    including incremental accuracy and average incremental accuracy \cite{castro2018end,hou2019learning,rebuffi2017icarl}. Additionally, we introduce phase accuracy. \textit{Incremental accuracy} \cite{castro2018end}, also known simply as accuracy in \cite{rebuffi2017icarl}, is the accuracy for classifying all the seen classes at the end of each training phase. It is the most commonly used metric but has the limitation of showing only \textcolor{black}{one single accuracy value} over the seen classes as a whole without giving any detail of how the model performs on separate groups of classes (learned incrementally in each phase). \textcolor{black}{\textit{Average incremental accuracy} simply takes the average of the incremental accuracy values obtained from the first training phase up to the current phase.} The results are given in parentheses for the last phase in Fig.~\ref{fig:cifar_and_imagenet_ac_a}.
    \textit{Phase accuracy} is evaluated at the end of the entire incremental training to present the average classification accuracy on separate groups of classes, where each group includes several classes that are added into the training at the same phase.
    It provides a breakdown look at whether the model would favor some groups of classes over the others as a consequence of catastrophic forgetting.

    \paragraph{{Implementation Details.}}
    For a fair comparison, we follow the baselines to use a 32-layer (respectively, 18-layer) ResNet \cite{he2016deep} as the feature extractor for CIFAR-100 (respectively, ImageNet), and a fully connected layer as the linear classifier. In particular, we remove the last ReLU layer for rectified cosine normalization. The memory size is fixed at 1000 or 2000, and the exemplars of old classes are chosen by the herd selection \cite{welling2009herding} proposed in iCaRL~\cite{rebuffi2017icarl}. At test time, we apply the same \emph{nearest-mean-of-exemplars} classification strategy as iCaRL.
    
\subsection{Incremental Accuracy Comparison}
\label{sec:Evaluation Resluts}

\begin{figure}[t]
    \centering
    \subfigure[$M$: 1K, FS]{
    \begin{minipage}[t]{0.235\textwidth}
    \label{fig:cifar_ac_a}
    \centering
    \includegraphics[width=1\textwidth]{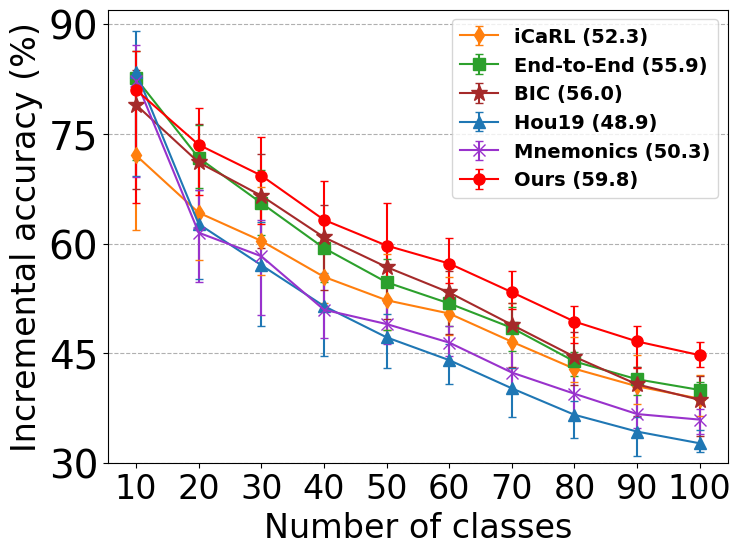}
    \end{minipage}%
    }%
    \subfigure[$M$: 2K, FS]{
    \begin{minipage}[t]{0.235\textwidth}
    \label{fig:cifar_ac_b}
    \centering
    \includegraphics[width=1\textwidth]{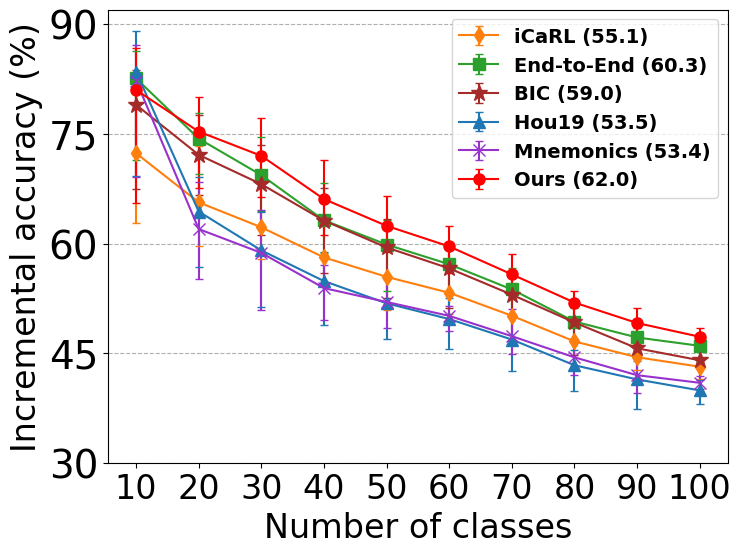}
    \end{minipage}%
    }%
    \subfigure[$M$: 1K, from 50]{
    \begin{minipage}[t]{0.235\textwidth}
    \label{fig:cifar_ac_c}
    \centering
    \includegraphics[width=1\textwidth]{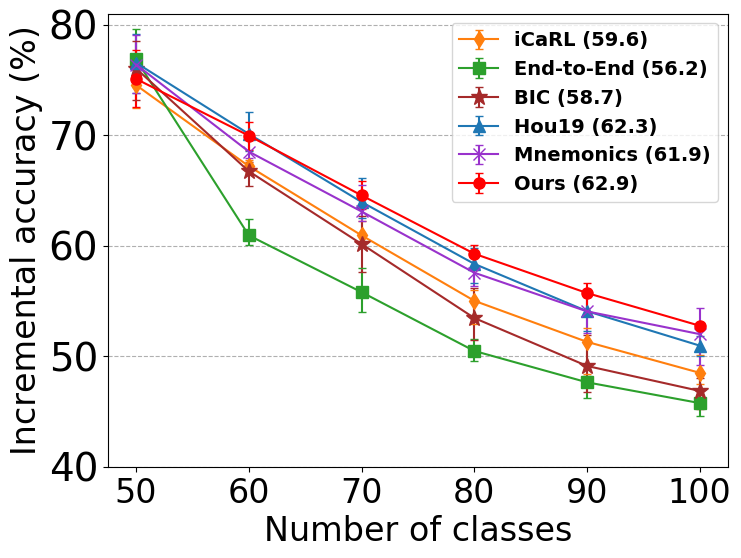}
    \end{minipage}
    }%
    \subfigure[$M$: 2K, from 50]{
    \begin{minipage}[t]{0.235\textwidth}
    \label{fig:cifar_ac_d}
    \centering
    \includegraphics[width=1\textwidth]{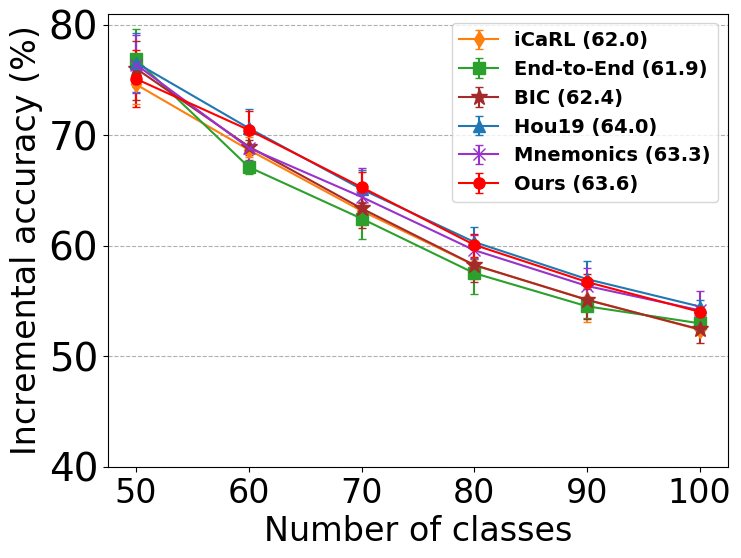}
    \end{minipage}
    }%
    \centering
    
    \subfigure[$M$: 1K, FS]{
    \begin{minipage}[t]{0.235\textwidth}
    \label{fig:imagenet_ac_a}
    \centering
    \includegraphics[width=1\textwidth]{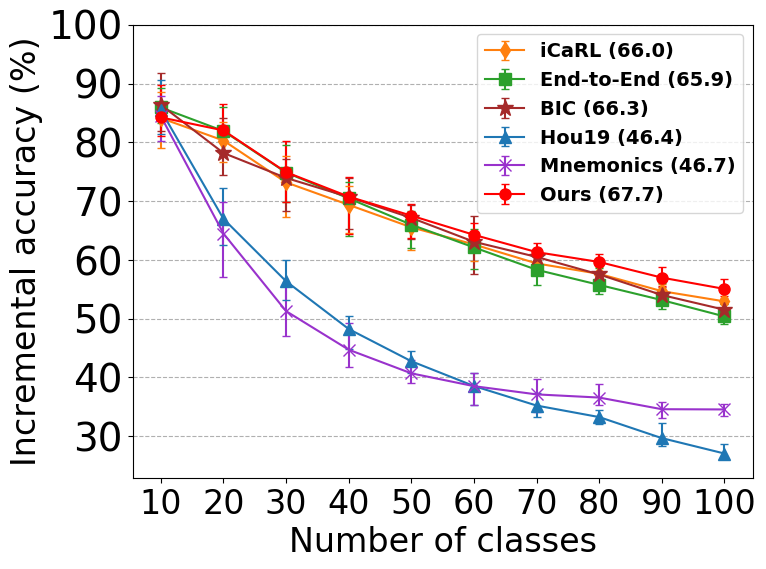}
    \end{minipage}%
    }%
    \subfigure[$M$: 2K, FS]{
    \begin{minipage}[t]{0.235\textwidth}
    \label{fig:imagenet_ac_b}
    \centering
    \includegraphics[width=1\textwidth]{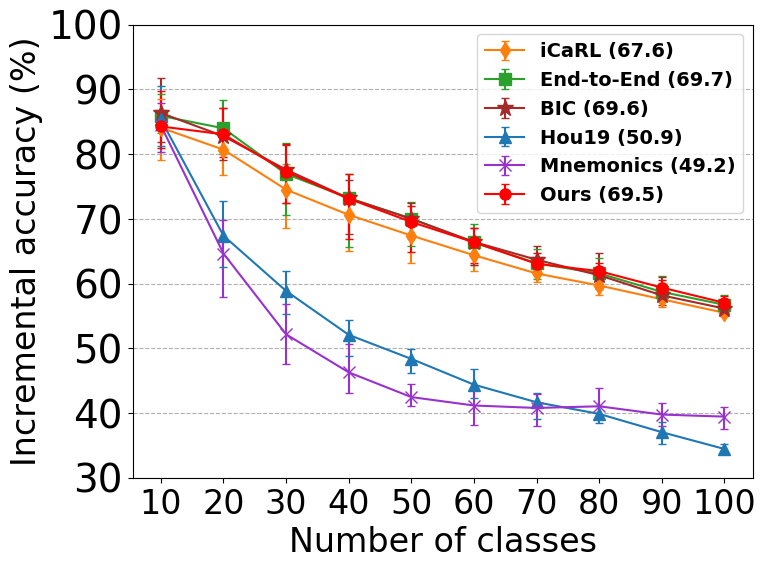}
    \end{minipage}%
    }%
    \subfigure[$M$: 1K, from 50]{
    \begin{minipage}[t]{0.235\textwidth}
    \label{fig:imagenet_ac_c}
    \centering
    \includegraphics[width=1\textwidth]{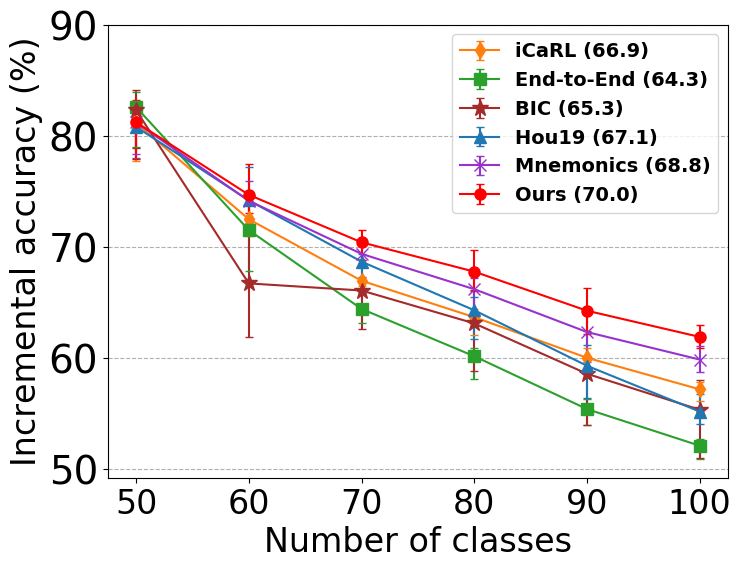}
    \end{minipage}
    }%
    \subfigure[$M$: 2K, from 50]{
    \begin{minipage}[t]{0.235\textwidth}
    \label{fig:imagenet_ac_d}
    \centering
    \includegraphics[width=1\textwidth]{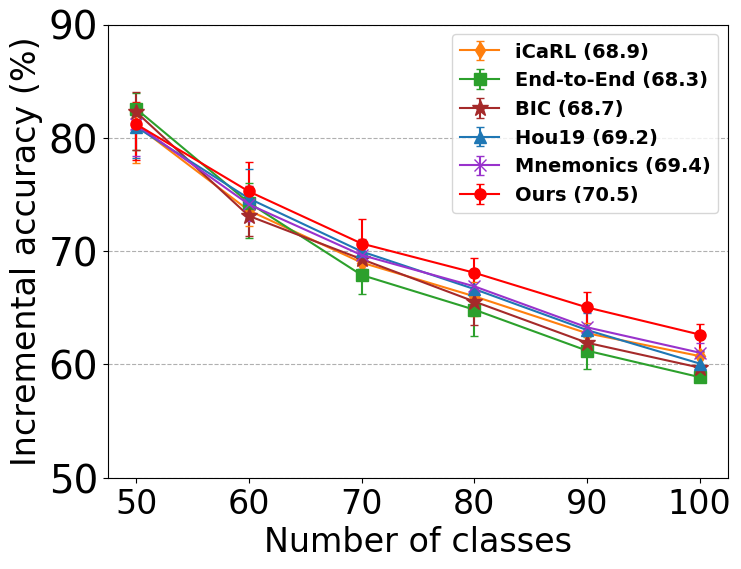}
    \end{minipage}
    }%
    \centering
    \caption{Incremental accuracy on CIFAR-100 (top row) and ImageNet (bottom row), for memory sizes $M=1K,2K$ and training scenarios ``FS" and ``from 50". Average incremental accuracy of each method is shown in parentheses.}
    \label{fig:cifar_and_imagenet_ac_a}
    \end{figure}

    Fig.~\ref{fig:cifar_and_imagenet_ac_a} presents incremental accuracy against the number of classes learned over the course of incremental learning, while Table~\ref{table:accuracy} summarizes the results at the end of the entire training. Notice that we use five random orderings of classes to build up class-incremental phases, where the metrics are evaluated and averaged over these five orderings. \textcolor{black}{Please also note that the figures showing the experimental results are better viewed in color and zoomed in for details.}

    We see that on CIFAR-100, when learning from scratch, our model outperforms all the baselines in almost every training phase, showing a significant boost in incremental accuracy especially with memory size 1000.
    In comparison, the recent state-of-the-arts from Hou19~\cite{hou2019learning} and Mnemonics~\cite{liu2020mnemonics} perform significantly worse as they rely heavily on a well-learned feature extractor (which is hard to obtain while being trained from scratch). When starting from 50 classes, ours performs the best with memory size 1000, and comparably to both Hou19~\cite{hou2019learning} and \textcolor{black}{Mnemonics~\cite{liu2020mnemonics}} with memory size 2000. Furthermore, our model shows higher average incremental accuracy than the baselines in nearly every setting, as shown in parentheses in Fig.~\ref{fig:cifar_and_imagenet_ac_a}.
    On ImageNet, our model performs comparably to \textcolor{black}{End-to-End~\cite{castro2018end} and BIC~\cite{wu2019large} when learning from scratch with $M=2K$} and is superior to the competing methods by a large margin when learning from 50 classes. 
    To sum up, our method shows consistently higher incremental accuracy than the other baselines on different datasets with varied characteristics.

    \begin{table}[t]
    \centering
    \renewcommand{\arraystretch}{1}
    \setlength{\tabcolsep}{5.0pt}
    \caption[Comparison of incremental accuracy at the end of training]{Comparison of incremental accuracy at the end of training}
    \label{table:accuracy}
    \begin{tabular}{c c c c c c c c c c c c}
    \hline
     Training scenario & \multicolumn{5}{c}{from scratch} && \multicolumn{5}{c}{from 50 classes}\\ \hline
     Dataset & \multicolumn{2}{c}{CIFAR} && \multicolumn{2}{c}{ImageNet}  && \multicolumn{2}{c}{CIFAR} && \multicolumn{2}{c}{ImageNet}\\ \hline
         Memory size & {1K} & {2K} && {1K} & {2K} && {1K} & {2K} && {1K} & {2K}  \\ \hline\hline
    iCaRL~\cite{rebuffi2017icarl}  & 38.8 & 43.1 && 52.9 & 55.5 && 48.5 & 52.4 && 57.2 & 60.7\\ \hline
    End-to-End~\cite{castro2018end}& 40.0 & 46.0 && 50.4 & 56.7 && 45.8 & 53.0 && 52.1 & 58.9\\ \hline
    BIC~\cite{wu2019large}  & 38.6 & 44.0 && 51.5 & 56.1  && 46.9 & 52.4 && 55.3 & 59.7\\ \hline
    Hou19~\cite{hou2019learning}  & 32.7 & 39.9 && 27.1 & 34.4 && 51.0 & \textbf{54.5} && 55.2 & 60.1\\ \hline
    Mnemonics~\cite{liu2020mnemonics} & 35.9 & 40.9 &&  34.6 & 39.4 && 52.0 & 54.2 && 59.8 & 61.0 \\ \hline
    Ours & \textbf{44.7} & \textbf{47.2} && \textbf{55.1} & \textbf{57.0} && \textbf{52.8} & 54.0 && \textbf{61.9} & \textbf{62.6} \\ \hline
    \end{tabular}
    \end{table}
    
\subsection{Phase Accuracy Comparison}
\label{sec:exp_balance_task} 
\begin{figure}[t]
    \centering
    \subfigure[$M$: 1K, from 50]{
    \begin{minipage}[t]{0.235\textwidth}
    \label{fig:task_a}
    \centering
    \includegraphics[width=1\textwidth]{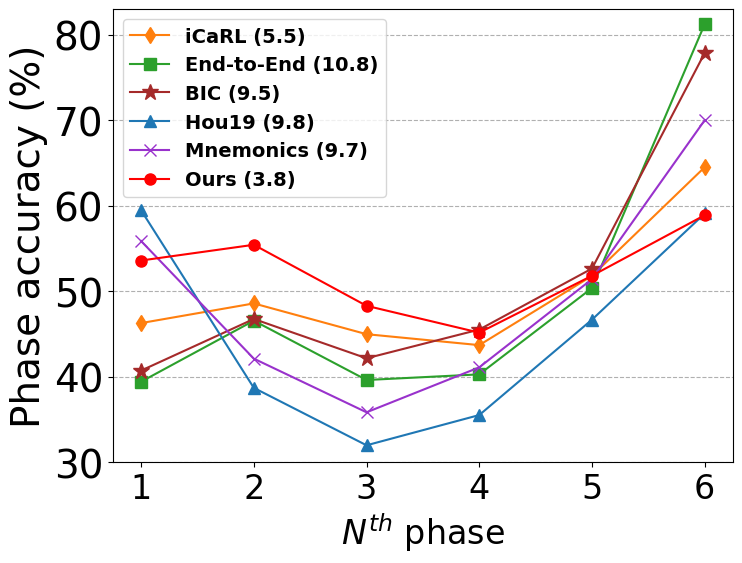}
    \end{minipage}%
    }%
    \subfigure[$M$: 2K, from 50]{
    \begin{minipage}[t]{0.235\textwidth}
    \label{fig:task_b}
    \centering
    \includegraphics[width=1\textwidth]{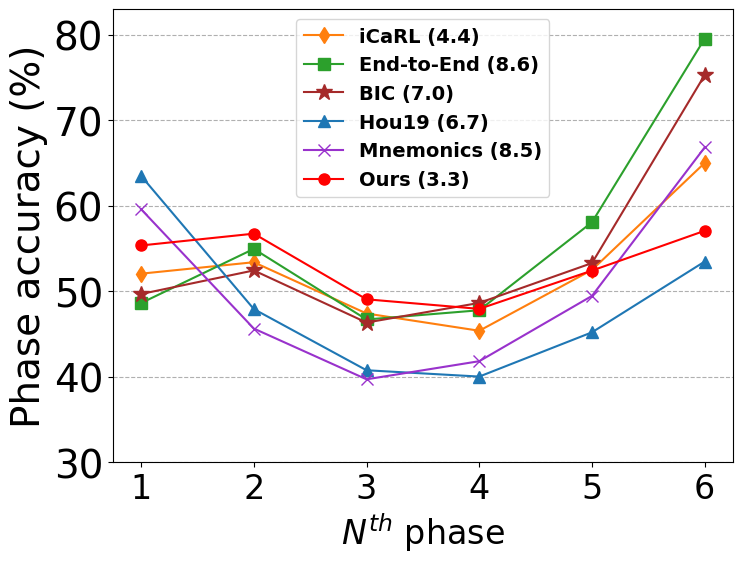}
    \end{minipage}%
    }%
    \subfigure[$M$: 1K, from 50]{
    \begin{minipage}[t]{0.235\textwidth}
    \label{fig:task_c}
    \centering
    \includegraphics[width=1\textwidth]{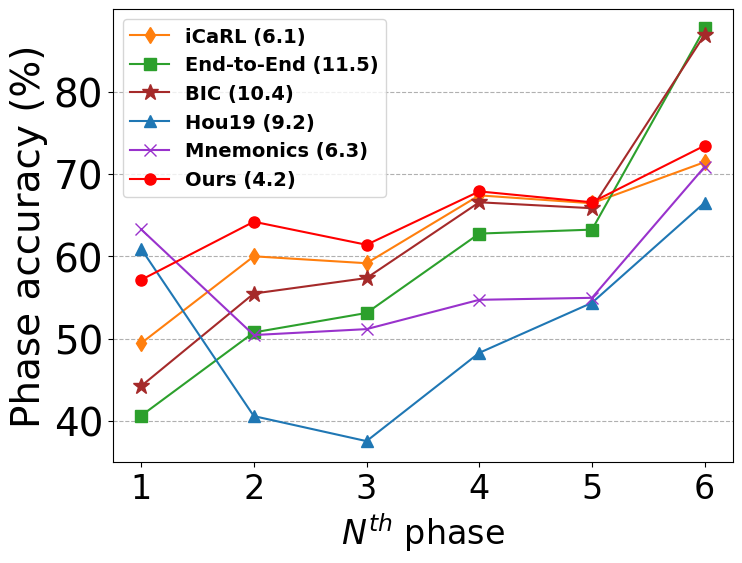}
    \end{minipage}
    }%
    \subfigure[$M$: 2K, from 50]{
    \begin{minipage}[t]{0.235\textwidth}
    \label{fig:task_d}
    \centering
    \includegraphics[width=1\textwidth]{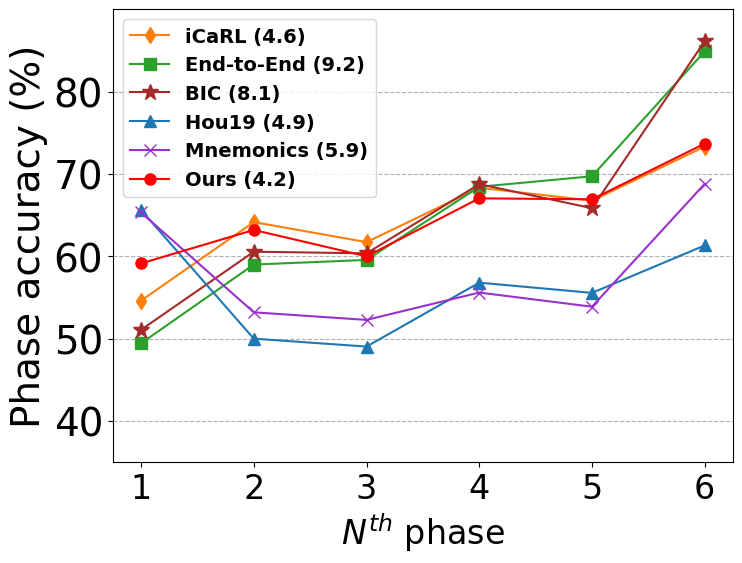}
    \end{minipage}
    }%
    \centering
    \caption{Phase accuracy comparison: (a)(b) are results on CIFAR-100 and (c)(d) on ImageNet. $M$ is the memory size, and the model is pre-trained with 50 classes. The mean absolute deviation in phase accuracy is shown in parentheses.}
    \label{fig:balance_accuracy}
    \end{figure}

    \textcolor{black}{
    Fig.~\ref{fig:balance_accuracy} presents the phase accuracy for different methods to compare their effectiveness in preserving knowledge of old classes. Generally, balanced phase accuracy is desirable. It is important to point out that there is a fundamental trade-off between incremental accuracy and phase accuracy. For a fair comparison, we particularly choose the training scenario ``from 50", where the baselines perform more closely to our method in terms of incremental accuracy evaluated at the end of the entire training. Shown in parentheses is the mean absolute deviation from the average of each method's phase accuracy. The smaller the deviation, the more balanced the classification accuracy is in different training phases.
    }
    \textcolor{black}{Our scheme is shown to achieve the minimum mean absolute deviation in phase accuracy on CIFAR-100 and ImageNet. 
    Remarkably, among all the baselines, Hou19~\cite{hou2019learning} and Mnemonics~\cite{liu2020mnemonics} have the closest incremental accuracy to ours, yet with considerable variations in phase accuracy.   
    }
    
\subsection{Ablation Studies}
\label{sec:ablation_rc_c}
\begin{figure}[t]
    \centering
    \subfigure[$M$: 1K, FS]{
    \begin{minipage}[t]{0.235\textwidth}
    \label{fig:ablation_cebce_in_1}
    \centering
    \includegraphics[width=1\textwidth]{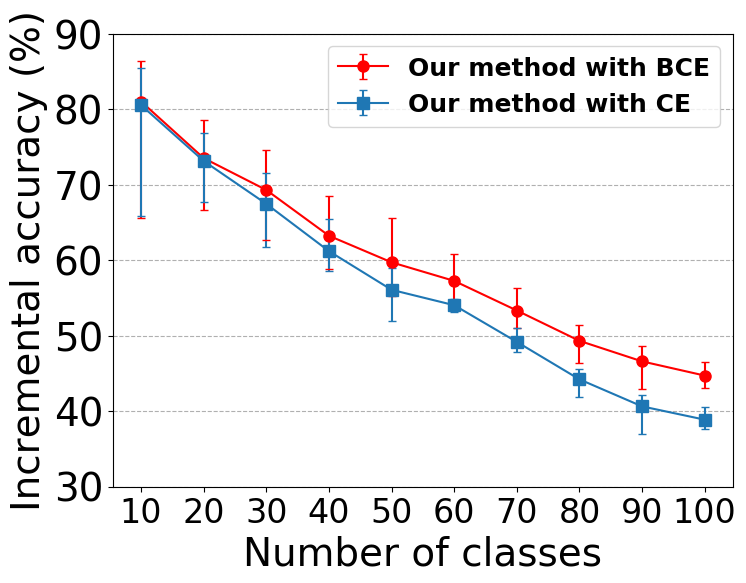}
    \end{minipage}%
    }%
    \subfigure[$M$: 2K, FS]{
    \begin{minipage}[t]{0.235\textwidth}
    \label{fig:ablation_cebce_in_2}
    \centering
    \includegraphics[width=1\textwidth]{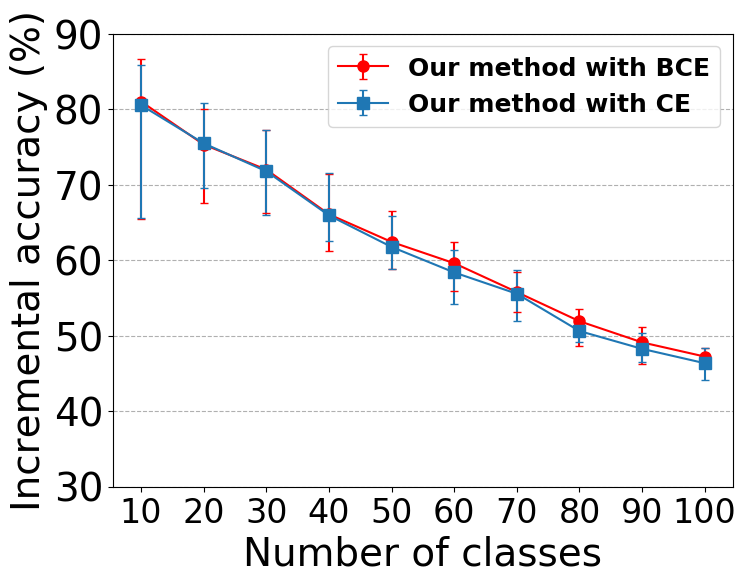}
    \end{minipage}%
    }%
    \subfigure[$M$: 1K, from 50]{
    \begin{minipage}[t]{0.235\textwidth}
    \label{fig:ablation_cebce_in_3}
    \centering
    \includegraphics[width=1\textwidth]{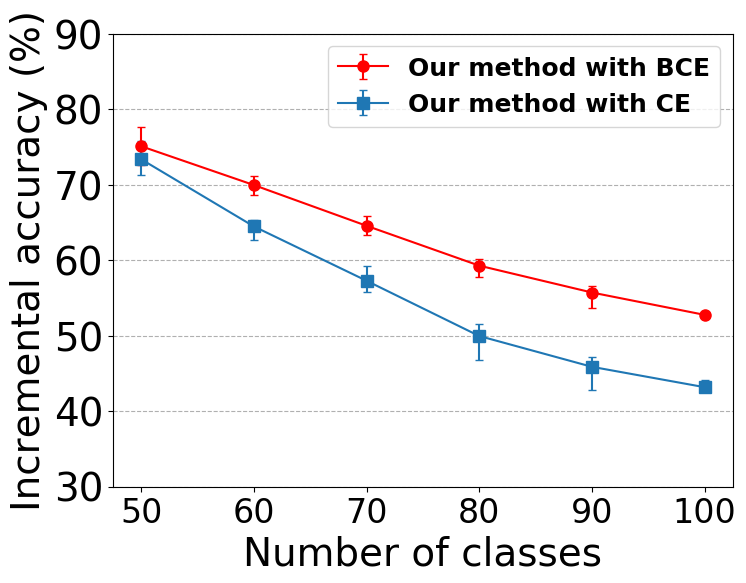}
    \end{minipage}
    }%
    \subfigure[$M$: 2K, from 50]{
    \begin{minipage}[t]{0.235\textwidth}
    \label{fig:ablation_cebce_in_4}
    \centering
    \includegraphics[width=1\textwidth]{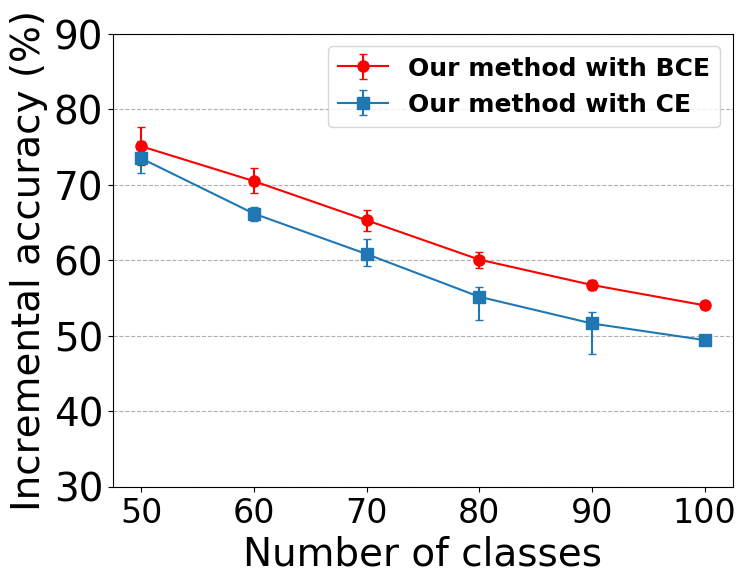}
    \end{minipage}
    }%
    
    \centering
    \subfigure[$M$: 1K, FS]{
    \begin{minipage}[t]{0.235\textwidth}
    \label{fig:ablation_uniform_cebce_in_1}
    \centering
    \includegraphics[width=1\textwidth]{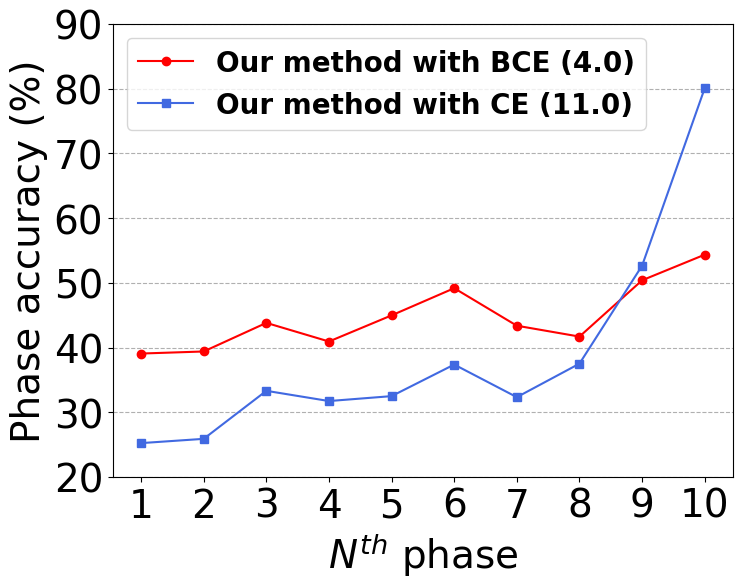}
    \end{minipage}%
    }%
    \subfigure[$M$: 2K, FS]{
    \begin{minipage}[t]{0.235\textwidth}
    \label{fig:ablation_uniform_cebce_in_2}
    \centering
    \includegraphics[width=1\textwidth]{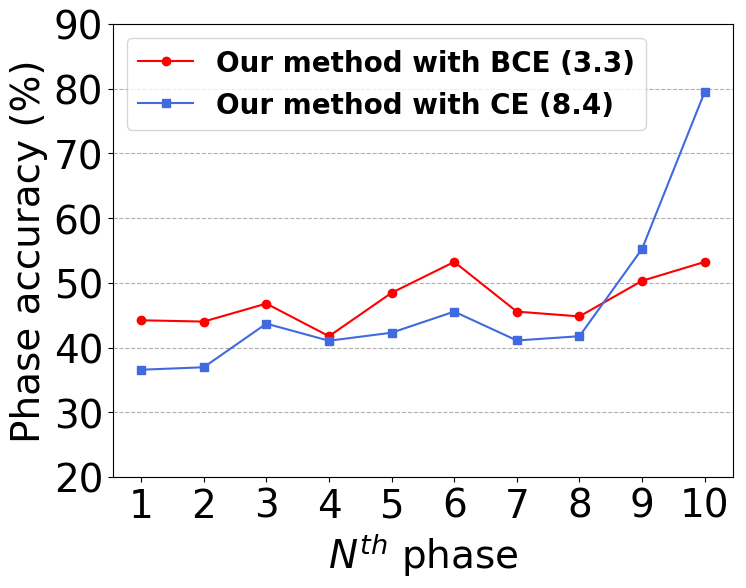}
    \end{minipage}%
    }%
    \subfigure[$M$: 1K, from 50]{
    \begin{minipage}[t]{0.235\textwidth}
    \label{fig:ablation_uniform_cebce_in_3}
    \centering
    \includegraphics[width=1\textwidth]{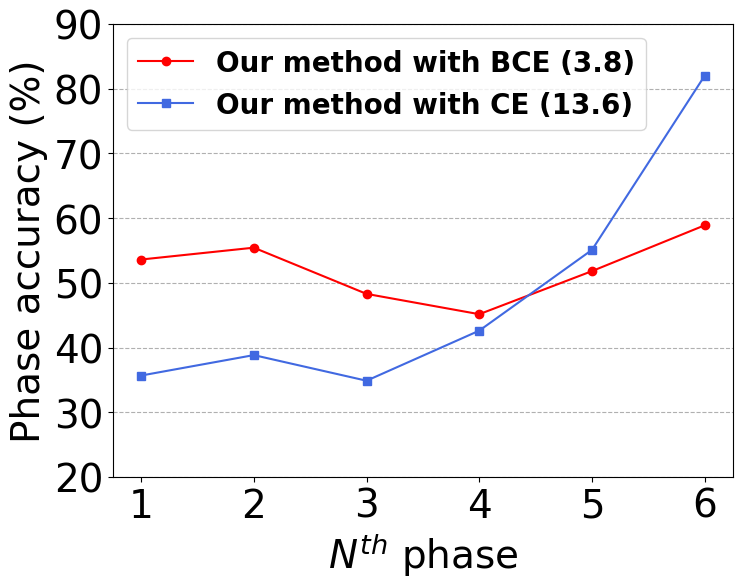}
    \end{minipage}
    }%
    \subfigure[$M$: 2K, from 50]{
    \begin{minipage}[t]{0.235\textwidth}
    \label{fig:ablation_uniform_cebce_in_4}
    \centering
    \includegraphics[width=1\textwidth]{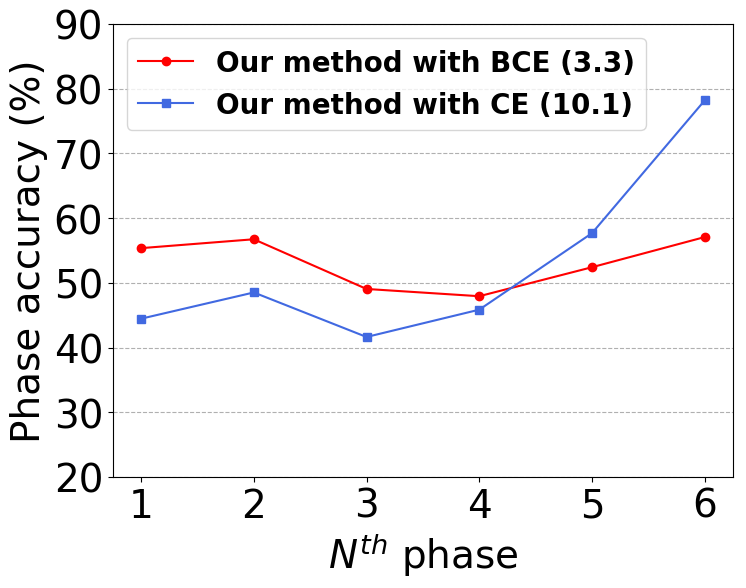}
    \end{minipage}
    }%
    \centering
    \caption{Comparison of binary cross-entropy and cross-entropy as the classification loss on CIFAR-100 for memory sizes $M = 1K, 2K$ and training scenarios ``FS" and ``from 50". The results are evaluated in terms of incremental accuracy. The bars are evaluated by five random orderings of classes.}
    \label{fig:ablation_cebce}
    \end{figure}

    \paragraph{\textbf{Binary Cross-entropy (BCE) versus Cross-entropy (CE) as Classification Loss.}} 
    We observe the performance differences by replacing BCE in our method with CE while keeping the other aspects (including rectified cosine normalization and weighted-Euclidean regularization) untouched.
    Fig.~\ref{fig:ablation_cebce} shows that BCE achieves consistently higher incremental accuracy than CE under all the settings. Moreover, BCE presents more evenly balanced phase accuracy whereas CE displays a tendency of biasing towards those new classes in the last phase.
    
\begin{figure}[t]
    \centering
    \subfigure[$M$: 1K, FS]{
    \begin{minipage}[t]{0.235\textwidth}
    \label{fig:ablation_rc_1}
    \centering
    \includegraphics[width=1\textwidth]{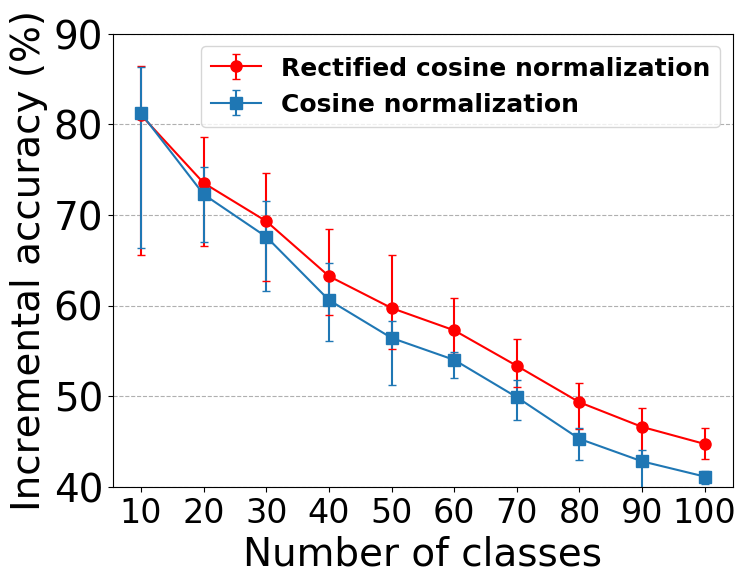}
    \end{minipage}%
    }%
    \subfigure[$M$: 2K, FS]{
    \begin{minipage}[t]{0.235\textwidth}
    \label{fig:ablation_rc_2}
    \centering
    \includegraphics[width=1\textwidth]{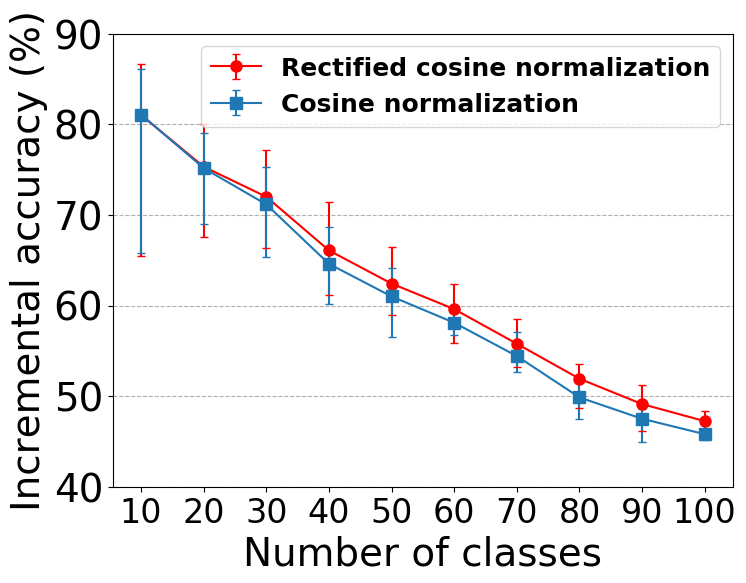}
    \end{minipage}%
    }%
    \subfigure[$M$: 1K, from 50]{
    \begin{minipage}[t]{0.235\textwidth}
    \label{fig:ablation_rc_3}
    \centering
    \includegraphics[width=1\textwidth]{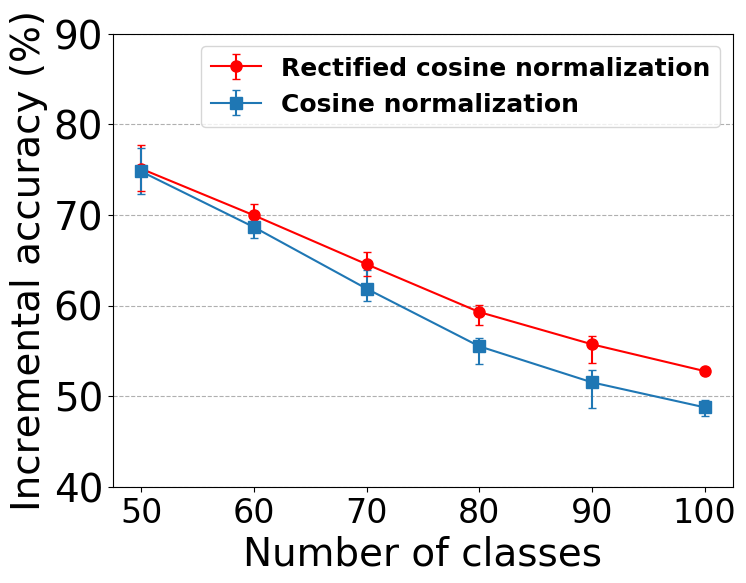}
    \end{minipage}
    }%
    \subfigure[$M$: 2K, from 50]{
    \begin{minipage}[t]{0.235\textwidth}
    \label{fig:ablation_rc_4}
    \centering
    \includegraphics[width=1\textwidth]{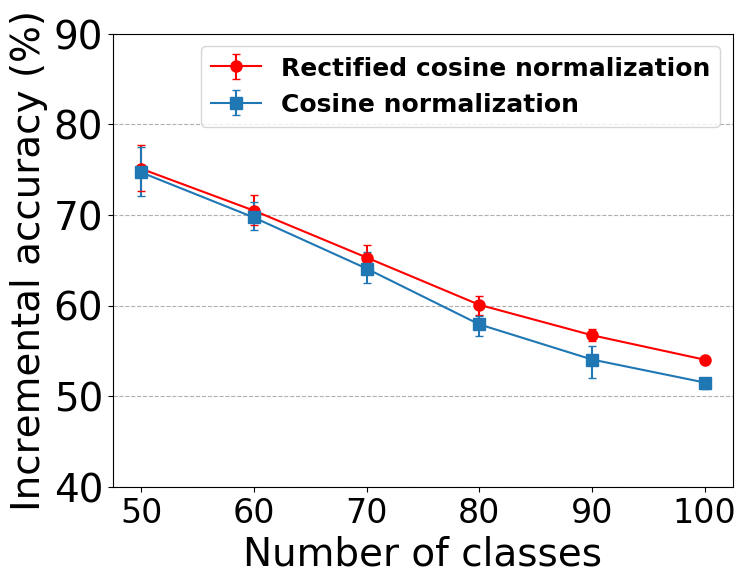}
    \end{minipage}
    }%
    \centering
    \caption{Comparison between rectified cosine normalization and cosine normalization on CIFAR-100 in terms of incremental accuracy, for memory sizes $M = 1K,2K$ and training scenarios ``FS" and ``from 50".}
    \label{fig:ablation_rc}
    \end{figure}

    \paragraph{\textbf{How effective is rectified cosine normalization in improving incremental accuracy?}}
    Fig.~\ref{fig:ablation_rc} presents results comparing rectified cosine normalization against cosine normalization in terms of incremental accuracy. Recall that rectified cosine normalization is proposed as a means to increase class separation and thus classification accuracy. To see the performance differences, we train a variant of our method by replacing solely rectified cosine normalization with cosine normalization (while keeping the other aspects the same). From Fig.~\ref{fig:ablation_rc}, our rectified cosine normalization starts with similar incremental accuracy to cosine normalization in early phases and gradually develops higher accuracy along the incremental learning process, which validates our analysis in Sec.~\ref{sec:classification}. 
    
\begin{figure}[t]
    \centering
    \subfigure[$M$: 1K, FS]{
    \begin{minipage}[t]{0.235\textwidth}
    \label{fig:ablation_dist_in_1}
    \centering
    \includegraphics[width=1\textwidth]{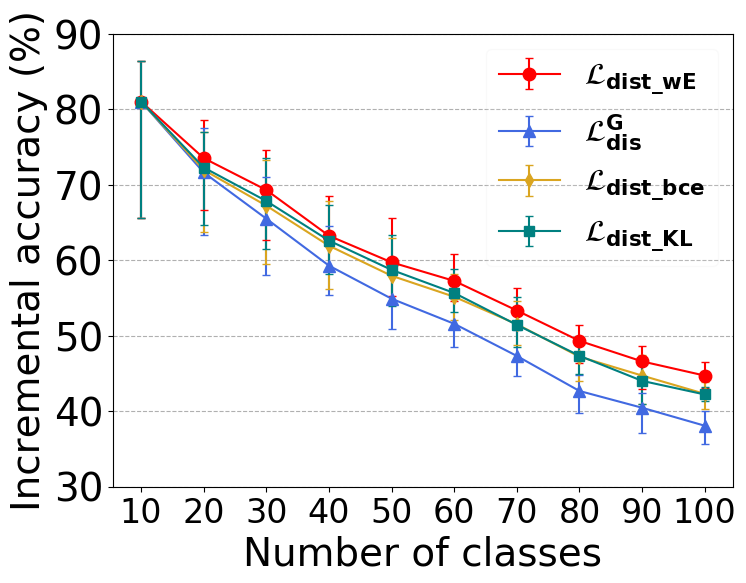}
    \end{minipage}%
    }%
    \subfigure[$M$: 2K, FS]{
    \begin{minipage}[t]{0.235\textwidth}
    \label{fig:ablation_dist_in_2}
    \centering
    \includegraphics[width=1\textwidth]{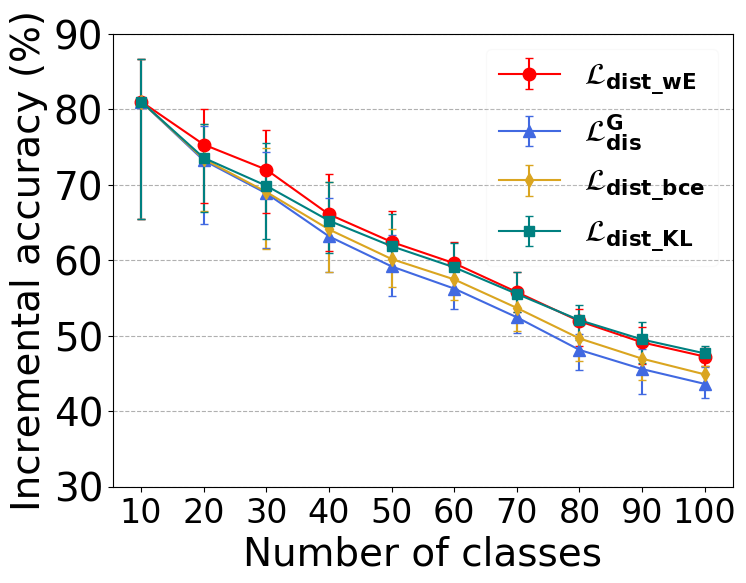}
    \end{minipage}%
    }%
    \subfigure[$M$: 1K, from 50]{
    \begin{minipage}[t]{0.235\textwidth}
    \label{fig:ablation_dist_in_3}
    \centering
    \includegraphics[width=1\textwidth]{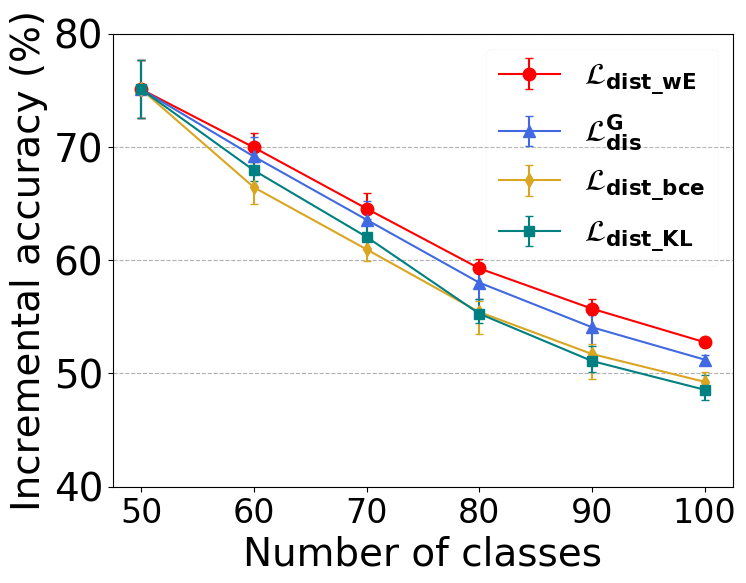}
    \end{minipage}
    }%
    \subfigure[$M$: 2K, from 50]{
    \begin{minipage}[t]{0.235\textwidth}
    \label{fig:ablation_dist_in_4}
    \centering
    \includegraphics[width=1\textwidth]{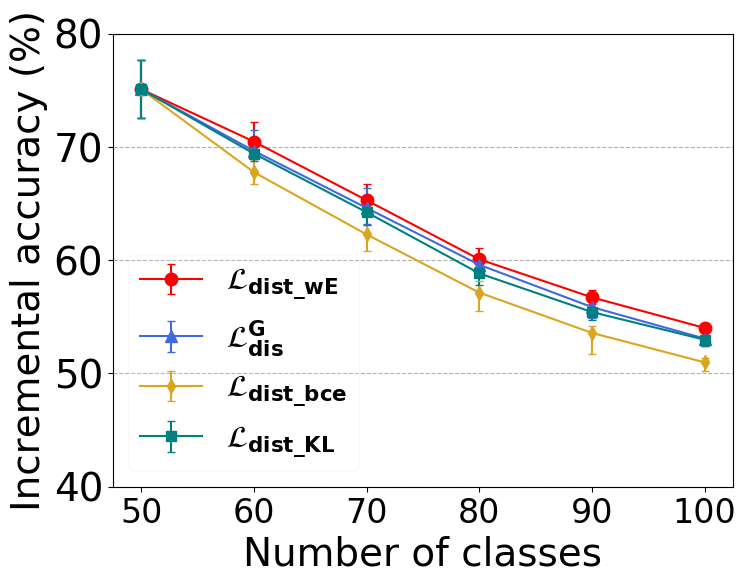}
    \end{minipage}
    }%
    
    \subfigure[$M$: 1K, FS]{
    \begin{minipage}[t]{0.235\textwidth}
    \label{fig:ablation_dist_1}
    \centering
    \includegraphics[width=1\textwidth]{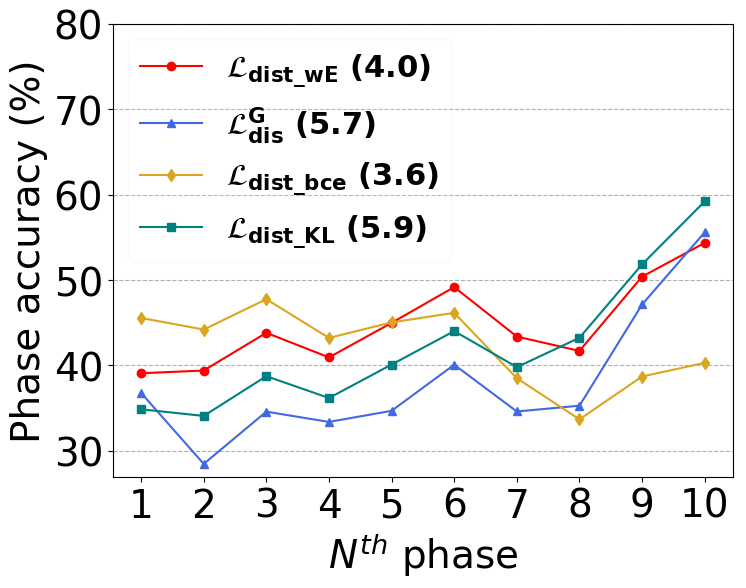}
    \end{minipage}%
    }%
    \subfigure[$M$: 2K, FS]{
    \begin{minipage}[t]{0.235\textwidth}
    \label{fig:ablation_dist_2}
    \centering
    \includegraphics[width=1\textwidth]{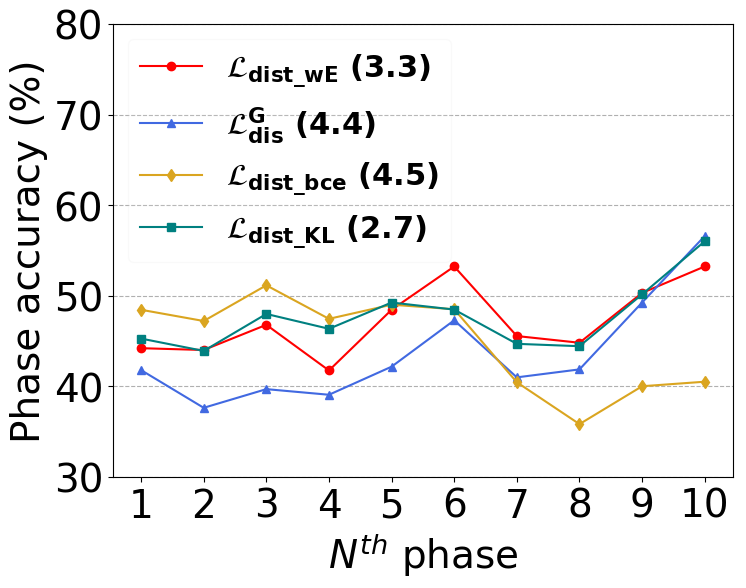}
    \end{minipage}%
    }%
    \subfigure[$M$: 1K, from 50]{
    \begin{minipage}[t]{0.235\textwidth}
    \label{fig:ablation_dist_3}
    \centering
    \includegraphics[width=1\textwidth]{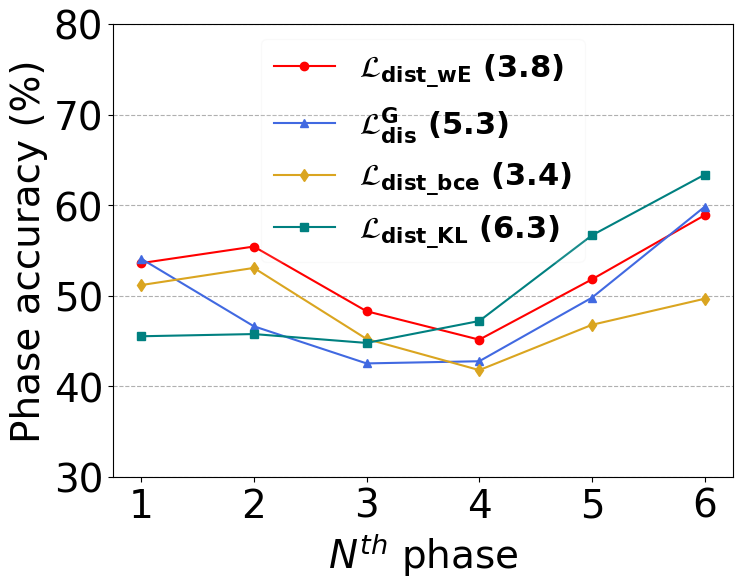}
    \end{minipage}
    }%
    \subfigure[$M$: 2K, from 50]{
    \begin{minipage}[t]{0.235\textwidth}
    \label{fig:ablation_dist_4}
    \centering
    \includegraphics[width=1\textwidth]{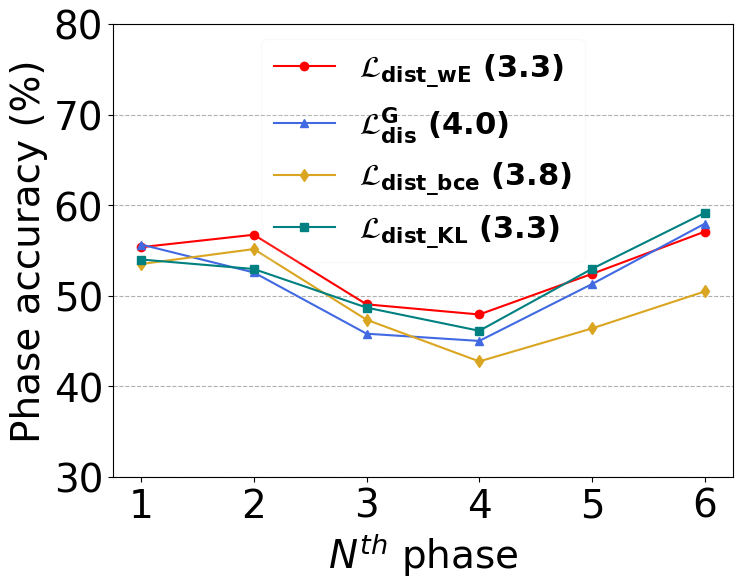}
    \end{minipage}
    }%
    \centering
    \caption{Comparison of various distillation losses on CIFAR-100, for memory sizes $M = 1K,2K$ and training scenarios ``FS" and ``from 50", with the mean absolute deviation in phase accuracy shown in parentheses.}
    \label{fig:ablation_dist}
    \end{figure}

    \paragraph{\textbf{Efficacy of weighted-Euclidean regularization in balancing new class learning and old knowledge preservation.}}
    Fig.~\ref{fig:ablation_dist} compares our weighted-Euclidean regularization $\mathcal{L}_{dist\_wE}$ with various distillation losses (see Sec.~\ref{sec:forgetting} for $\mathcal{L}_{dist\_bce}$ and $\mathcal{L}_{dist\_KL}$, and  Hou19~\cite{hou2019learning} for ${L}^{G}_{dis}$) by examining their trade-offs between incremental and phase accuracy. 
    These variants are tested on the same base model of using BCE and rectified cosine normalization.
    We see that weighted-Euclidean regularization achieves higher incremental accuracy among the competing methods, and shows lowest or the second-lowest mean absolute deviation (MAD) in phase accuracy in most cases (cf. training all classes together in batch mode leads to $2.5$ MAD in phase accuracy).
    These suggest that our weighted-Euclidean regularization enables the model to strike a better balance between new class learning and old knowledge preservation. 
    
\begin{figure}[h!]
    \centering
    \subfigure[$M$: 1K, FS]{
    \begin{minipage}[t]{0.235\textwidth}
    \label{fig:ablation_alpha_in_1}
    \centering
    \includegraphics[width=1\textwidth]{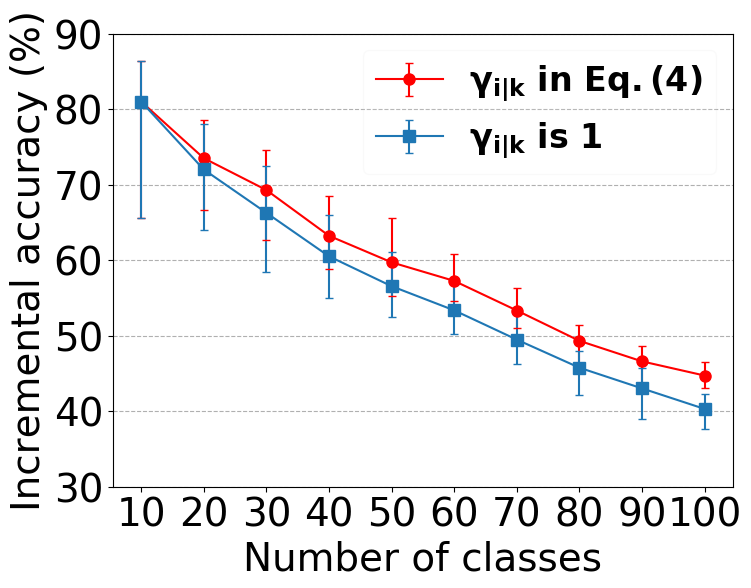}
    \end{minipage}%
    }%
    \subfigure[$M$: 2K, FS]{
    \begin{minipage}[t]{0.235\textwidth}
    \label{fig:ablation_alpha_in_2}
    \centering
    \includegraphics[width=1\textwidth]{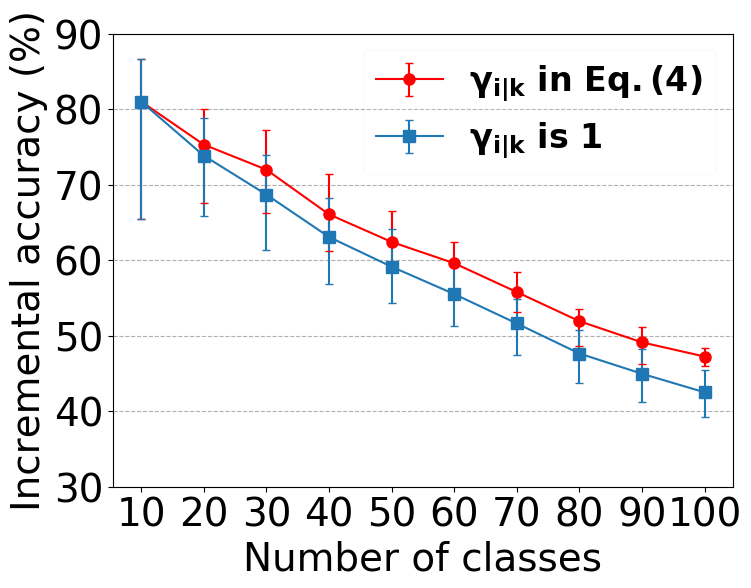}
    \end{minipage}%
    }%
    \subfigure[$M$: 1K, from 50]{
    \begin{minipage}[t]{0.235\textwidth}
    \label{fig:ablation_alpha_in_3}
    \centering
    \includegraphics[width=1\textwidth]{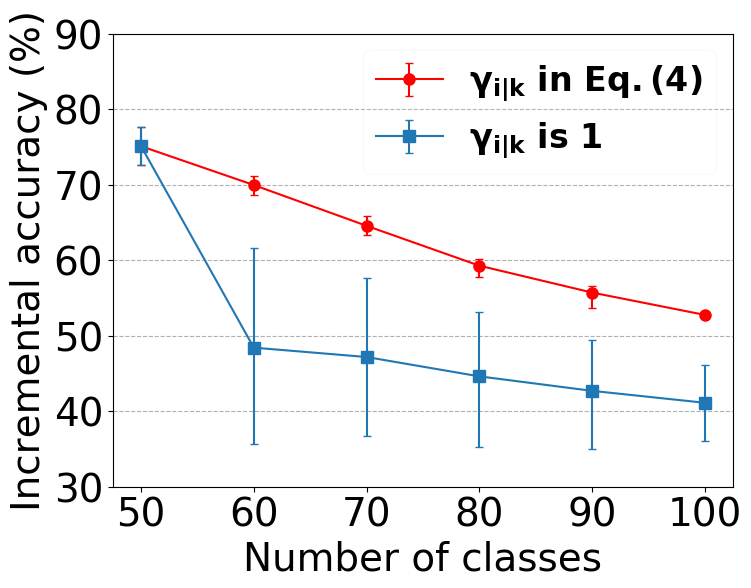}
    \end{minipage}
    }%
    \subfigure[$M$: 2K, from 50]{
    \begin{minipage}[t]{0.235\textwidth}
    \label{fig:ablation_alpha_in_4}
    \centering
    \includegraphics[width=1\textwidth]{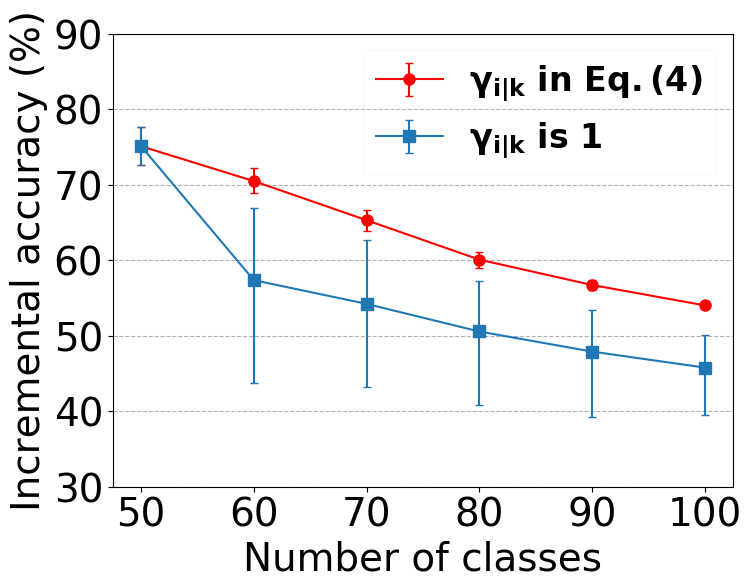}
    \end{minipage}
    }%
    
    \subfigure[$M$: 1K, FS]{
    \begin{minipage}[t]{0.235\textwidth}
    \label{fig:ablation_alpha_1}
    \centering
    \includegraphics[width=1\textwidth]{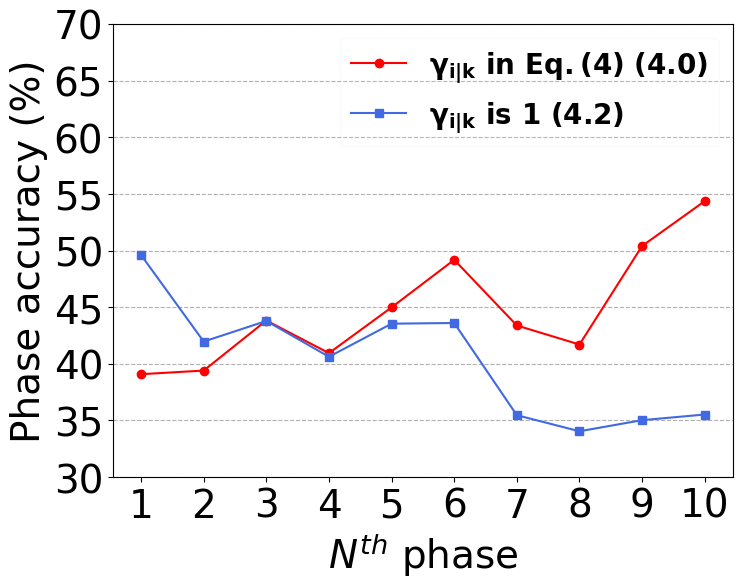}
    \end{minipage}%
    }%
    \subfigure[$M$: 2K, FS]{
    \begin{minipage}[t]{0.235\textwidth}
    \label{fig:ablation_alpha_2}
    \centering
    \includegraphics[width=1\textwidth]{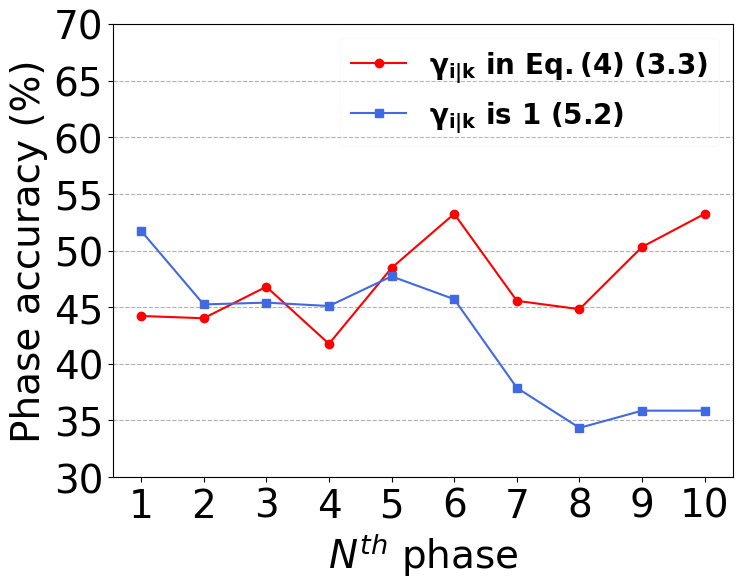}
    \end{minipage}%
    }%
    \subfigure[$M$: 1K, from 50]{
    \begin{minipage}[t]{0.235\textwidth}
    \label{fig:ablation_alpha_3}
    \centering
    \includegraphics[width=1\textwidth]{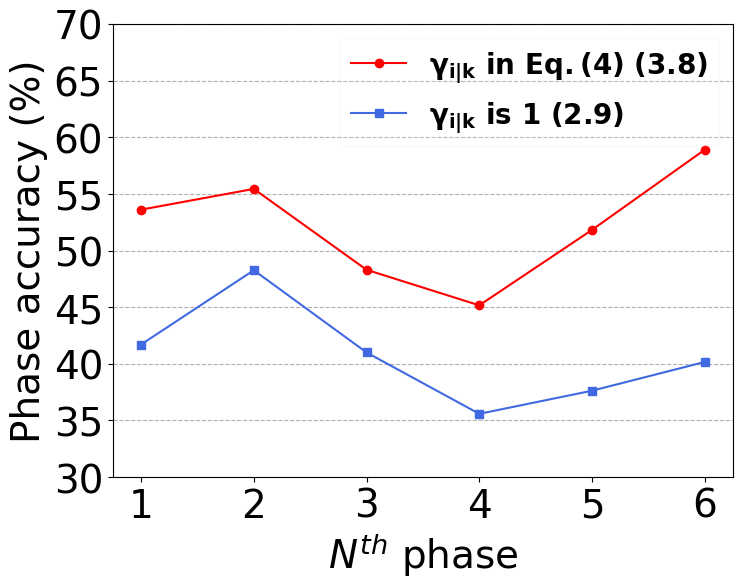}
    \end{minipage}
    }%
    \subfigure[$M$: 2K, from 50]{
    \begin{minipage}[t]{0.235\textwidth}
    \label{fig:ablation_alpha_4}
    \centering
    \includegraphics[width=1\textwidth]{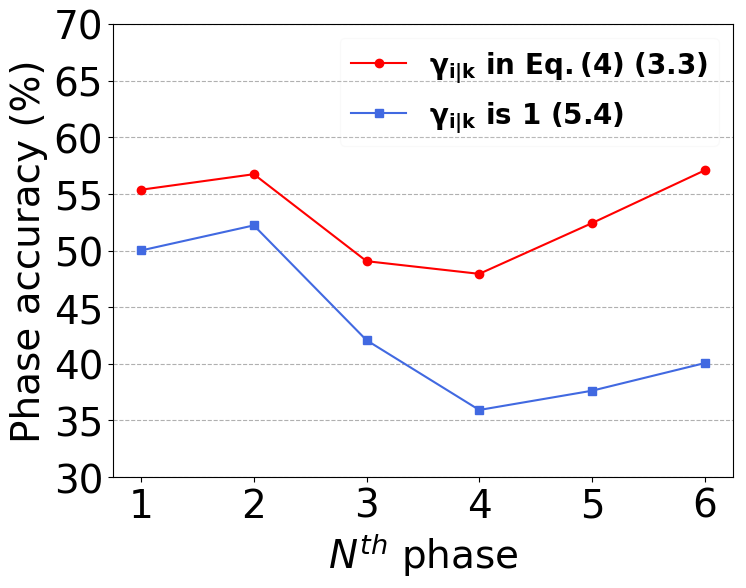}
    \end{minipage}
    }%
    \centering
    \caption{Comparison of uniform versus non-uniform $\gamma_{i|k}$ assignment on CIFAR-100 for memory sizes $M =1K, 2K $ and training scenario ``FS" and ``from 50," with the mean absolute deviation in phase accuracy (the bottom row) shown in parentheses. The bars in the top row are evaluated by five random orderings of classes.}
    \label{fig:ablation_alpha}
    \end{figure}

    \paragraph{\textbf{Uniform vs. Non-uniform $\gamma_{i|k}$ Assignment.}}
    This experiment investigates the sole benefits of prioritized edge preservation (i.e. non-uniform assignment of $\gamma_{i|k}$) in our weighted-Euclidean regularization. We construct a variant of our method by setting $\gamma_{i|k}$ in Eq.~\eqref{equ:euclidean_factor} to $1$ and observe the performance differences. As shown in Fig.~\ref{fig:ablation_alpha}, the non-uniform assignment of $\gamma_{i|k}$ in our scheme leads to higher incremental accuracy and lower mean absolute deviations in phase accuracy. In addition, the smaller variations in incremental accuracy (cp. the bars in the top row of Fig.~\ref{fig:ablation_alpha}) suggest that our method with prioritized edge preservation is less sensitive to the learning order of object classes.
    
%% file: conclusion.tex
\section{Conclusion}
\label{sec:conclusion}
    We introduce a feature-graph preservation perspective to study the essence of various knowledge distillation techniques in how they preserve the structure of the feature graph along the phases of class-incremental learning. Developed based on three desirable properties, our weighted-Euclidean regularization is shown to benefit the balance between new class learning and old knowledge preservation. Furthermore, our rectified cosine normalization effectively improves incremental accuracy due to greater class separation. With all the techniques combined, our model consistently outperforms the state-of-the-art baselines on CIFAR-100 and ImageNet datasets.

%% file: appendix.tex
\begin{figure}[t]
\centering
\includegraphics[width=1.\textwidth]{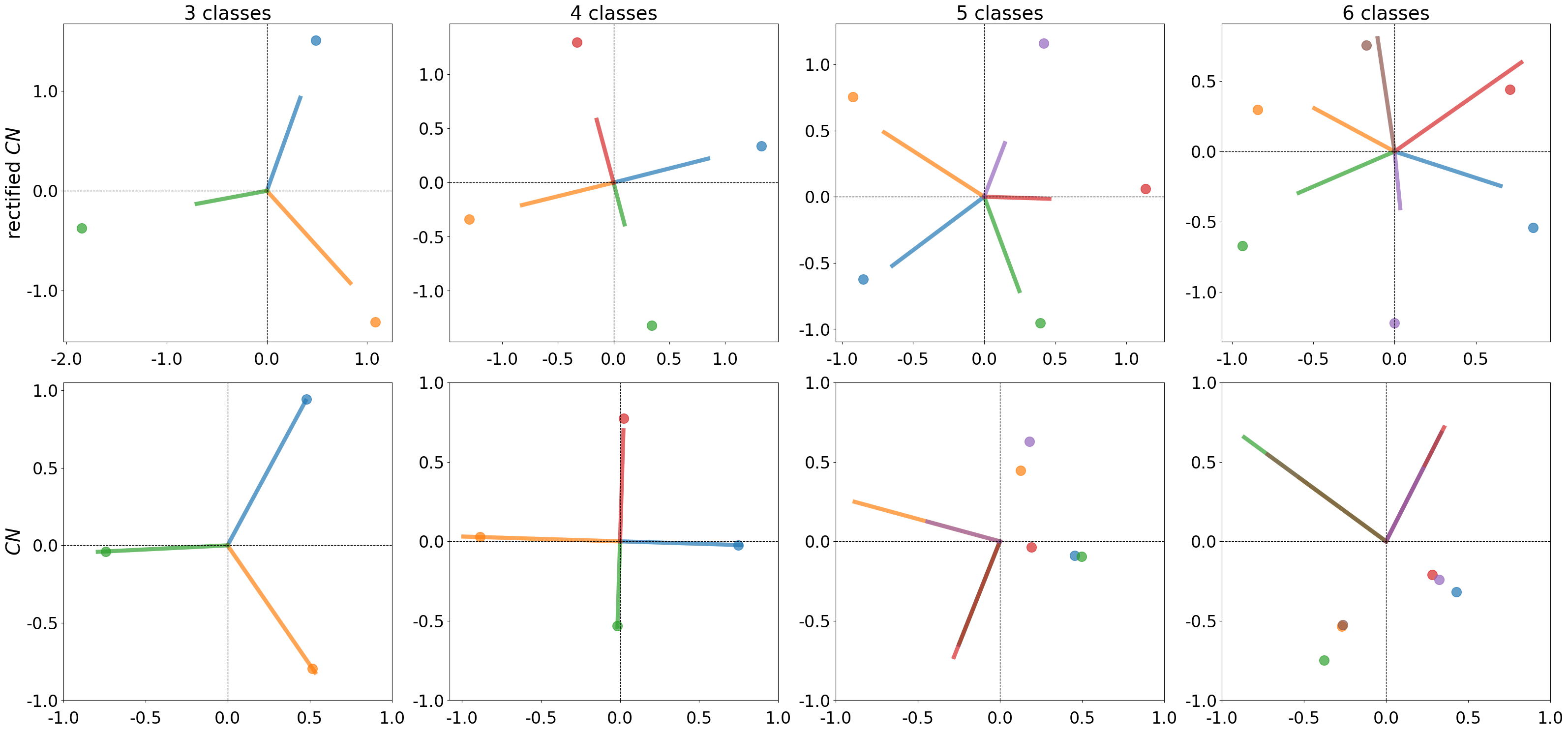}
\caption{Comparison of our rectified cosine normalization (denoted as $\text{rectified-}CN$) and the typical cosine normalization (denoted as $CN$) with different number of classes. Dots in various colors represent the learned feature vector $f_k$ for class $k$, and the line segment of the same color denotes the corresponding weight vector $w_k$.}
\label{fig:BCE_example}
\end{figure}

\section{Rectified Cosine Normalization}
This section conducts a simulation to justify our rectified cosine normalization. It is reported empirically in the main manuscript that using rectified cosine normalization for training can better encourage the separation between classes.

Specifically, in evaluating the binary cross-entropy loss $\mathcal{L}_{bce}(x_k)$, our rectified cosine normalization computes the activation ${a}_{i|k}$ of an image $x_k$ for class $i$ to be ${a}_{i|k} = \bar{W}^{T}_{i}\bar{F_k}$, while the cosine normalization in \cite{hou2019learning} uses ${a}_{i|k} = \bar{w}^{T}_{i}\bar{f_k}$. The bar indicates $l_{2}$-normalization, $W_i=(w_i,b_i)$ is formed by the concatenation of the classification weight vector $w_i$ and an additional learnable bias $b_i$; likewise, $F_k=(f_k,1)$ (referred to as the augmented feature representation) concatenates the feature $f_k$ (referred to as the original feature representation) of the image $x_k$ and a constant bias $1$. Due to the $l_{2}$-normalization, the activation ${a}_{i|k}$ is in the range of $[-1,1]$. We thus introduce a learnable $\eta$ to control the curvature of the sigmoid function $\sigma(\cdot)$:
\begin{equation}
\sigma (a_{i|k}) = \frac{1}{1+exp(-\eta {a}_{i|k})}.
\label{activation}
\end{equation}

Fig.~\ref{fig:BCE_example} illustrates the benefits of our rectified cosine normalization, taking the 2-dimensional cases as examples. In these examples, each class has only one trainable data $f_k$ (or $F_k$) and the corresponding class embedding $w_k$ (or $W_k$), where $f_k,w_k \in R^2$ and $F_k,W_k \in R^3$. They are learned via the steepest descent algorithm. Notably, $f_k$ (or $F_k$) is considered a parameter to be solved, with the aim of simulating the ideal case where the feature extractor has enough capacity.   
The first row presents the results of training with rectified cosine normalization in the augmented feature space $F_k=(f_k,1)$ with $f_k=(x,y) \in R^2$. Both the learned features $F_k$ and the class embeddings $W_i=(w_i,b_i)$ are projected onto the $xy$-plane for visualization and comparison. The second row corresponds to the results of using typical cosine normalization, for which both $f_k$ and $w_i$ are learned directly in the original feature space without augmentation.
It is clear to see that in the cases with 3 and 4 classes, both rectified cosine normalization and typical cosine normalization work well. The features $f_k$ point in the same direction as their class embeddings $w_k$, as expected. We note that minimizing the $\mathcal{L}_{bce}(x_k)$ (equation (6) in the main manuscript) requires that (1) $F_k$ should be separated from the class embeddings $W_i$ not of the same class (i.e. $i \neq k$) by at least 90 degrees and that (2) the angle between $F_k$ and $W_k$ of the same class should be smaller than 90 degrees and preferably be 0 degrees. These apply to $f_k$ and $w_i$ learned directly in the original feature space. When the number of classes is small, these requirements are easy to fulfill, be it in the 2-dimensional original feature space or in the 3-dimensional augmented feature space. They however become difficult to hold simultaneously in the 2-dimensional feature space when the number of classes increases beyond 4. Intriguingly, in those cases, the feature vectors of different classes may collapse into few modes, putting much emphasis on the first requirement due mainly to the excessive amount of negative examples (i.e. for a given $f_k$, the number of class embeddings $w_i,i \neq k$ not of the same class is much larger than the case $i=k$). In contrast, both requirements can still hold in the 3-dimensional augmented feature space. We observe that this is achieved by setting all $b_i$'s to negative values and having $F_k$ and $W_k$ (respectively, $W_i,i \neq k$) point in approximately similar directions (respectively, in significantly different directions) when projected onto the xy-plane. The larger degree of freedom in adapting the class embeddings $W_i$ to the two requirements in the higher dimensional augmented feature space explains the generally better separation between feature vectors $f_k$ with our rectified cosine normalization in the lower dimensional original feature space. 

\section{Further Training Details}
This section provides further training details. For a fair comparison, we follow the settings in iCaRL~\cite{rebuffi2017icarl}. On CIFAR-100, the learning rate starts from 2.0 and is divided by 5 after 49 and 63 epochs. Moreover, 70 epochs are run in each incremental training phase. Likewise, on ImageNet the learning rate starts from 2.0 and is divided by 5 after 20, 30, 40 and 50 epochs. Each training phase has 60 epochs. On both datasets, the optimizer is SGD and the batch size is 128.

\section{Detail Formulation of Our Evaluation}
We follow the common metrics in prior works \cite{rebuffi2017icarl,castro2018end,hou2019learning} to evaluate the performance of incremental learning; that is, incremental accuracy, average incremental accuracy. Furthermore, we additionally propose a metric called \textit{phase accuracy}. The details of these metrics are described as follows.

\paragraph{\textbf{Incremental Accuracy.}}
Let $A_{j,i}$ be the average accuracy of classes learned in phase $i$ after training the network sequentially to phase $j$, where $i < j$. Incremental accuracy is defined as $A_j = \sum_{i=1}^{j} A_{j,i}$, which is the accuracy for classifying all the seen classes at the end of training phase $j$. It is the most commonly-used metric, but has the limitation of showing only the accuracy over the seen classes as a whole without giving any detail of how the model performs on separate groups of classes (learned incrementally in each phase).

\paragraph{\textbf{Average Incremental Accuracy.}}
Average incremental accuracy accumulates the incremental accuracy, obtained from each training phase, up to the current phase $j$ and then takes the average, i.e. $\frac{1}{j}\sum_{i=1}^{j}A_{i}$.

\paragraph{\textbf{Phase Accuracy.}}
A model that learns well incrementally should present a balanced distribution of $A_{j,i}, i \in \{1,...,j\}$. Thus, we propose phase accuracy to evaluate at the end of the entire incremental training to present the classification accuracy on separate groups of classes (a group represents a phase in this thesis). It provides a breakdown look at whether the model would favor some groups of classes over the others as a result of catastrophic forgetting.

\section{More Ablation Experiments}

\subsection{Symmetric vs. Asymmetric Discrepancy Measure $\mathcal{D}(p^*_{i|k},p_{i|k})$}
\begin{figure}[t]
    \centering
    \subfigure[$M$: 1K, FS]{
    \begin{minipage}[t]{0.235\textwidth}
    \label{fig:ablation_webce_in_1}
    \centering
    \includegraphics[width=1\textwidth]{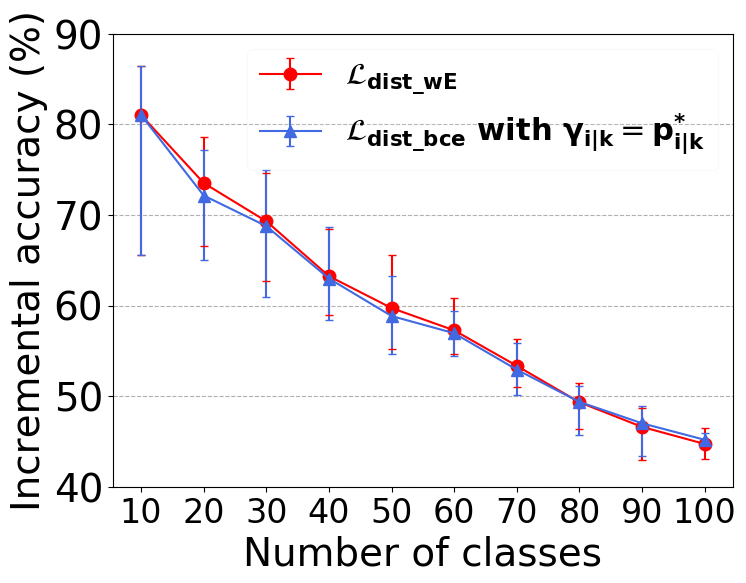}
    \end{minipage}%
    }%
    \subfigure[$M$: 2K, FS]{
    \begin{minipage}[t]{0.235\textwidth}
    \label{fig:ablation_webce_in_2}
    \centering
    \includegraphics[width=1\textwidth]{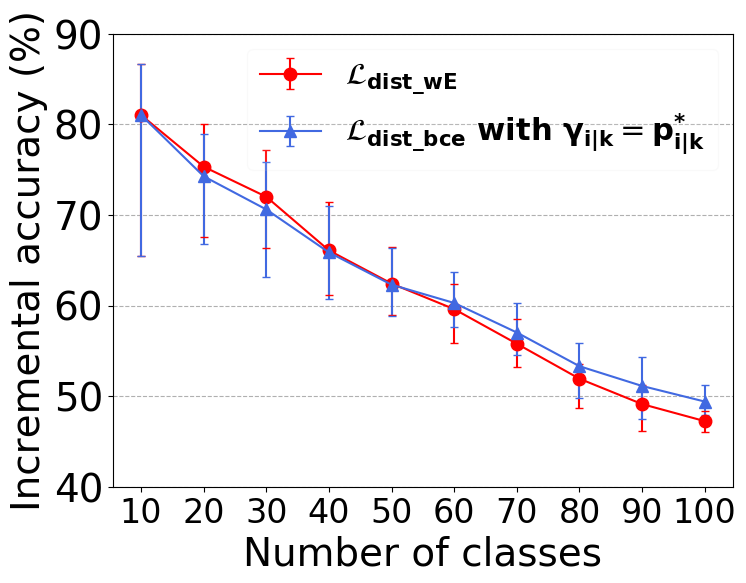}
    \end{minipage}%
    }%
    \subfigure[$M$: 1K, from 50]{
    \begin{minipage}[t]{0.235\textwidth}
    \label{fig:ablation_webce_in_3}
    \centering
    \includegraphics[width=1\textwidth]{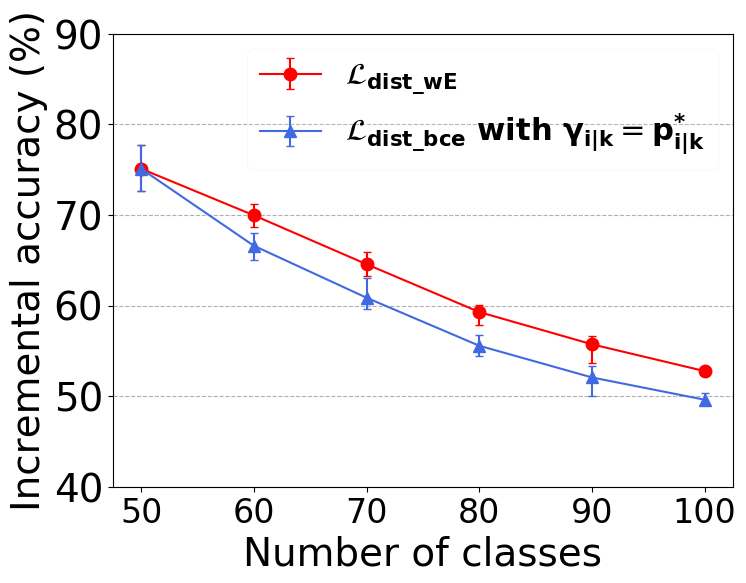}
    \end{minipage}
    }%
    \subfigure[$M$: 2K, from 50]{
    \begin{minipage}[t]{0.235\textwidth}
    \label{fig:ablation_webce_in_4}
    \centering
    \includegraphics[width=1\textwidth]{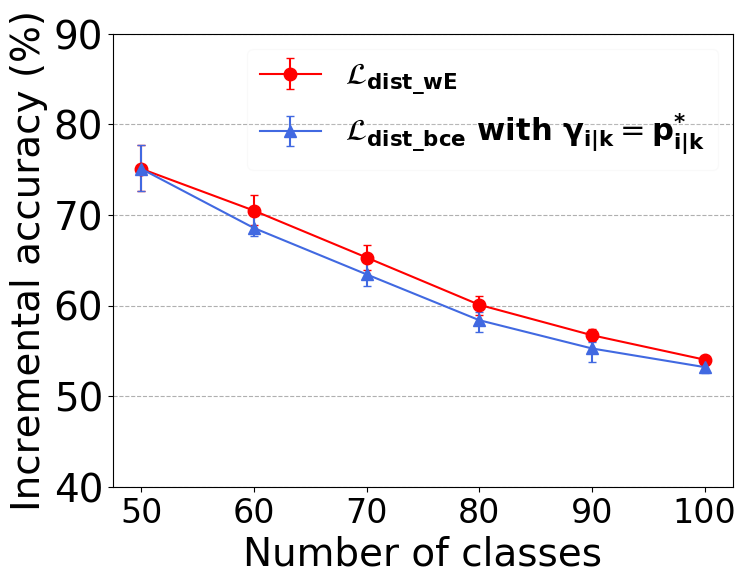}
    \end{minipage}
    }%
    
    %\vspace{-3mm}
    \subfigure[$M$: 1K, FS]{
    \begin{minipage}[t]{0.235\textwidth}
    \label{fig:ablation_webce_1}
    \centering
    \includegraphics[width=1\textwidth]{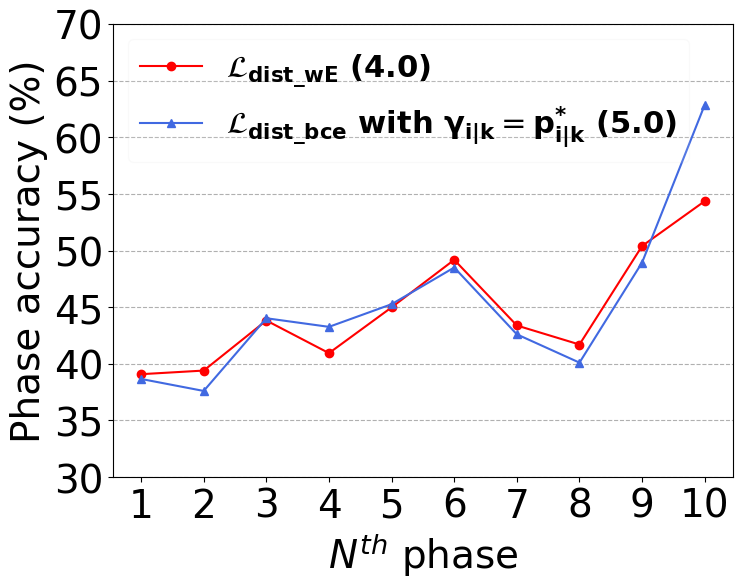}
    \end{minipage}%
    }%
    \subfigure[$M$: 2K, FS]{
    \begin{minipage}[t]{0.235\textwidth}
    \label{fig:ablation_webce_2}
    \centering
    \includegraphics[width=1\textwidth]{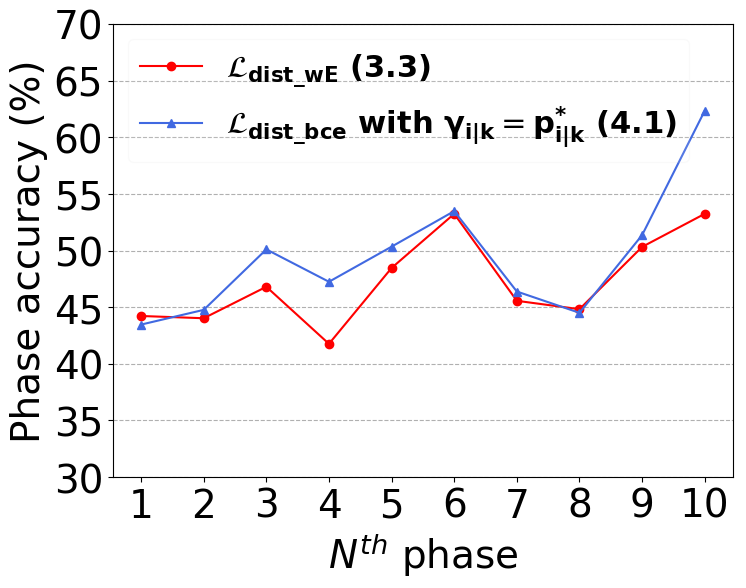}
    \end{minipage}%
    }%
    \subfigure[$M$: 1K, from 50]{
    \begin{minipage}[t]{0.235\textwidth}
    \label{fig:ablation_webce_3}
    \centering
    \includegraphics[width=1\textwidth]{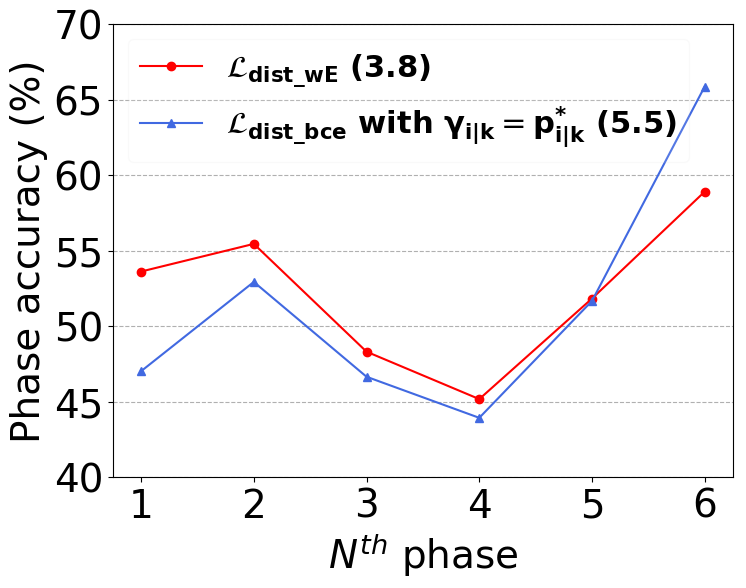}
    \end{minipage}
    }%
    \subfigure[$M$: 2K, from 50]{
    \begin{minipage}[t]{0.235\textwidth}
    \label{fig:ablation_webce_4}
    \centering
    \includegraphics[width=1\textwidth]{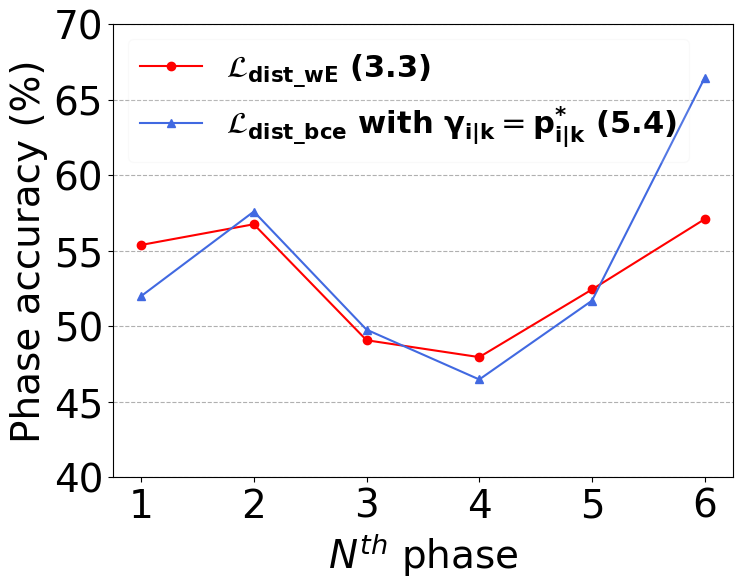}
    \end{minipage}
    }%
    \centering
    %\vspace{-3mm}
    \caption{Comparison of discrepancy measure $\mathcal{D}(p^*_{i|k},p_{i|k})$ on CIFAR-100 for memory sizes $M = 1K, 2K$ and training scenario ``FS" and ``from 50," with the mean absolute deviation in phase accuracy shown in parentheses. The bars in the top row are evaluated by five random orderings of classes.}
    %\vspace{-4mm}
    \label{fig:ablation_webce}
    \end{figure}
    
    We argue in Sec. 3.1 of the main manuscript that an $a^*_{i|k}$-symmetric discrepancy measure is desirable. That is, $\mathcal{D}(p^*_{i|k},p_{i|k})$ is preferably symmetric with respect to $a^*_{i|k}$ when viewed as a function of $a_{i|k}$. To single out its benefits, we change our distillation loss $\mathcal{L}_{dist\_wE}$ by replacing the $\mathcal{D}(p^*_{i|k},p_{i|k})$ in Eq.~(3) with the binary cross-entropy (BCE) between $p^*_{i|k}$ and $p_{i|k}$, and setting $\gamma_{i|k}=p^*_{i|k}$, while keeping the other design aspects (including BCE as classification loss and rectified cosine normalization) untouched. Note that these changes lead to a scheme (denoted collectively as $\mathcal{L}_{dist\_bce}$ with our $\gamma_{i|k}$) that differs from ours only in adopting an asymmetric discrepancy measure with respect to $a^*_{i|k}$. From the top row of Fig.~\ref{fig:ablation_webce}, it is interesting to see that the replacement of $\mathcal{D}(p^*_{i|k},p_{i|k})=(\log p^*_{i|k} - \log p_{i|k})^2$ with BCE does not show much impact on incremental accuracy. However, our $a^*_{i|k}$-symmetric discrepancy measure  presents more balanced phase accuracy, achieving smaller mean absolute deviations (cp. the bottom row of Fig.~\ref{fig:ablation_webce}). This highlights the fact that it allows our model to strike a better balance between incremental accuracy and phase accuracy. 
    
\subsection{iCaRL with vs. without Rectified Cosine Normalization}

    \begin{figure}[t]
    \centering
    \subfigure[$M$: 1K, FS]{
    \begin{minipage}[t]{0.235\textwidth}
    \label{fig:ablation_norm_icarl_in_1}
    \centering
    \includegraphics[width=1\textwidth]{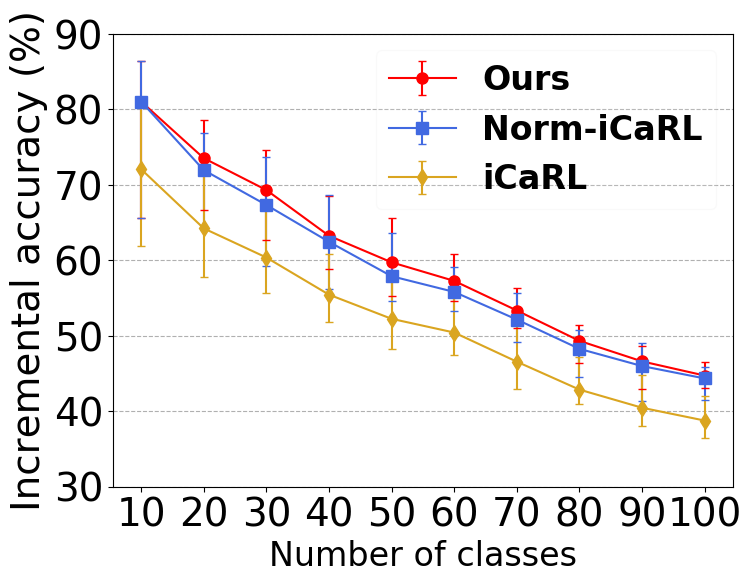}
    \end{minipage}%
    }%
    \subfigure[$M$: 2K, FS]{
    \begin{minipage}[t]{0.235\textwidth}
    \label{fig:ablation_norm_icarl_in_2}
    \centering
    \includegraphics[width=1\textwidth]{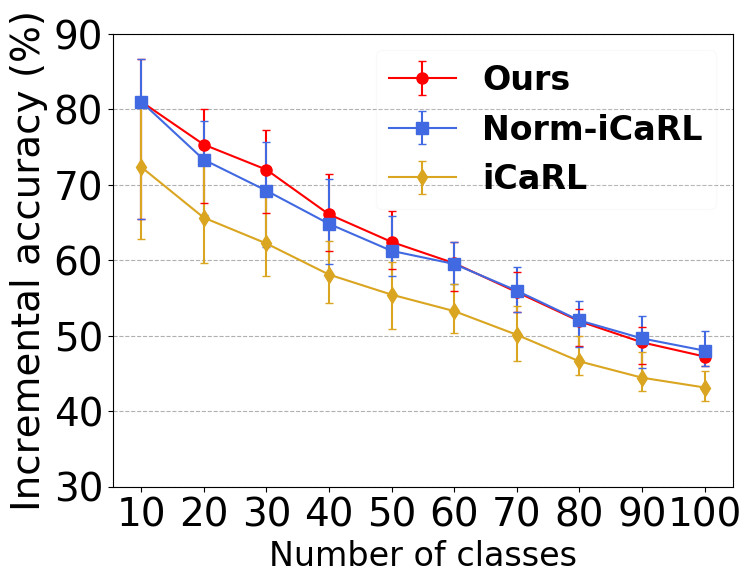}
    \end{minipage}%
    }%
    \subfigure[$M$: 1K, from 50]{
    \begin{minipage}[t]{0.235\textwidth}
    \label{fig:ablation_norm_icarl_in_3}
    \centering
    \includegraphics[width=1\textwidth]{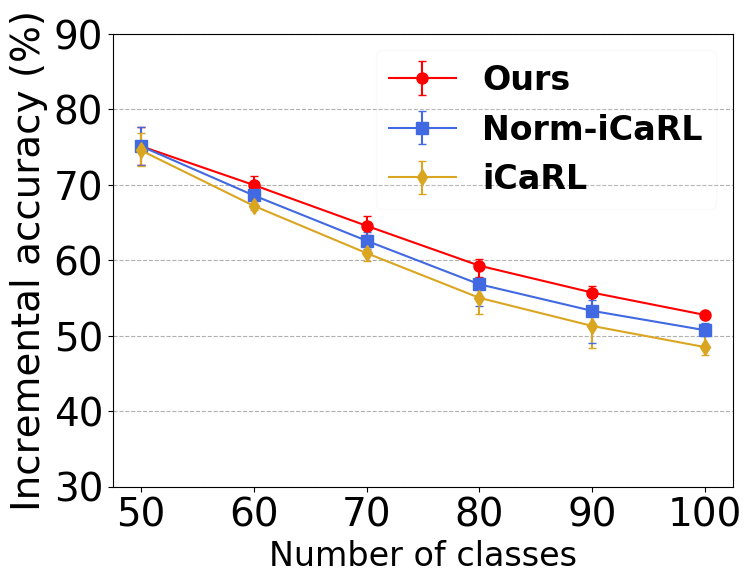}
    \end{minipage}
    }%
    \subfigure[$M$: 2K, from 50]{
    \begin{minipage}[t]{0.235\textwidth}
    \label{fig:ablation_norm_icarl_in_4}
    \centering
    \includegraphics[width=1\textwidth]{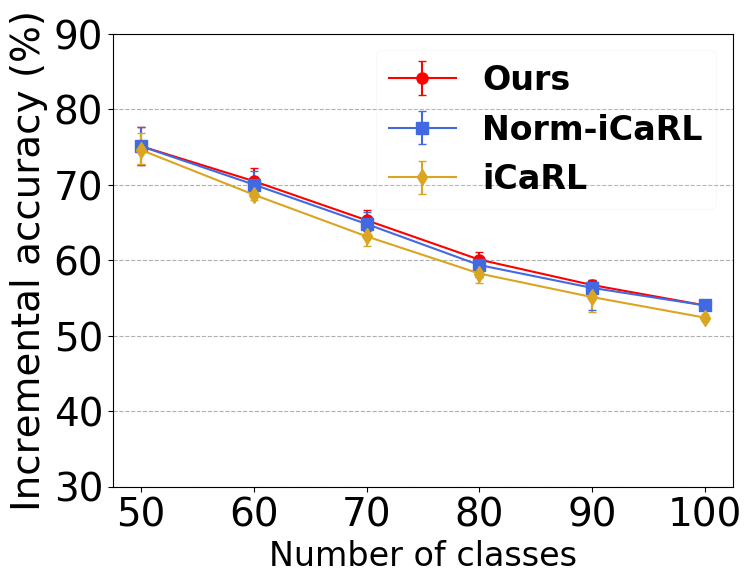}
    \end{minipage}
    }%
    
    %\vspace{-3mm}
    \subfigure[$M$: 1K, FS]{
    \begin{minipage}[t]{0.235\textwidth}
    \label{fig:ablation_norm_icarl_1}
    \centering
    \includegraphics[width=1\textwidth]{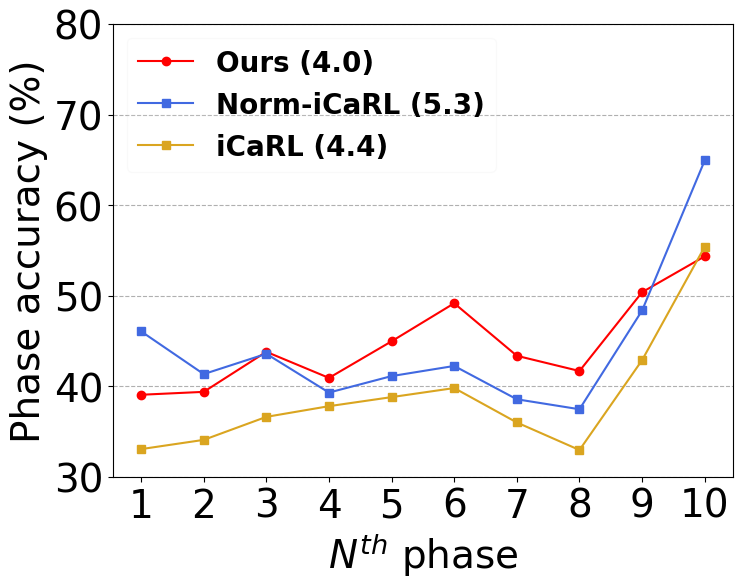}
    \end{minipage}%
    }%
    \subfigure[$M$: 2K, FS]{
    \begin{minipage}[t]{0.235\textwidth}
    \label{fig:ablation_norm_icarl_2}
    \centering
    \includegraphics[width=1\textwidth]{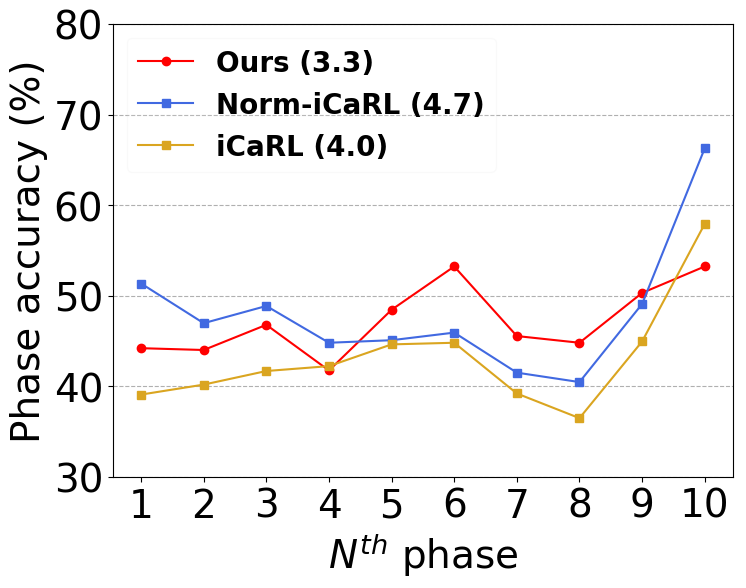}
    \end{minipage}%
    }%
    \subfigure[$M$: 1K, from 50]{
    \begin{minipage}[t]{0.235\textwidth}
    \label{fig:ablation_norm_icarl_3}
    \centering
    \includegraphics[width=1\textwidth]{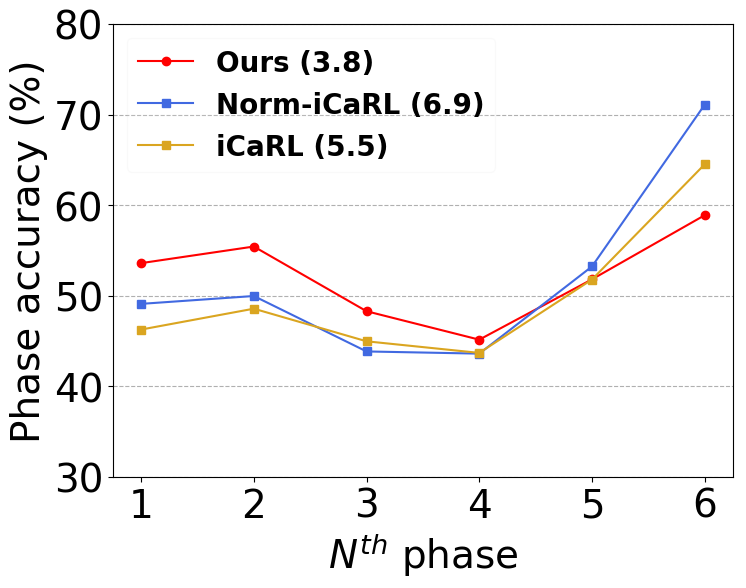}
    \end{minipage}
    }%
    \subfigure[$M$: 2K, from 50]{
    \begin{minipage}[t]{0.235\textwidth}
    \label{fig:ablation_norm_icarl_4}
    \centering
    \includegraphics[width=1\textwidth]{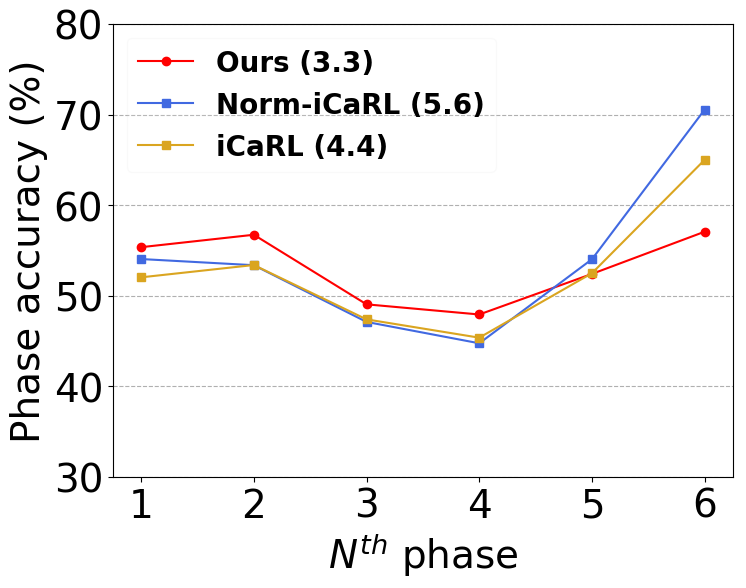}
    \end{minipage}
    }%
    \centering
    %\vspace{-3mm}
    \caption{Comparison of iCaRL with/without rectified cosine normalization and our method on CIFAR-100 for memory sizes $M =1K,2K$ and training scenarios ''FS" and ''from 50," with the mean absolute deviation in phase accuracy shown in parentheses. The bars in the top row are evaluated by five random orderings of classes.}
    %\vspace{-4mm}
    \label{fig:ablation_norm_icarl}
    \end{figure}
    
    This section investigates whether our rectified cosine normalization could also benefit iCaRL~\cite{rebuffi2017icarl}. Compared in Fig.~\ref{fig:ablation_norm_icarl} are iCaRL~\cite{rebuffi2017icarl} with (denoted as Norm-iCaRL) and without (denoted as iCaRL) rectified cosine normalization. Note that the other design aspects of iCaRL remain the same. As a benchmark, our scheme is also presented. A comparison between Norm-iCaRL and iCaRL in the top row of Fig.~\ref{fig:ablation_norm_icarl} reveals that rectified cosine normalization does help improve the incremental accuracy of iCaRL~\cite{rebuffi2017icarl}, reducing largely the performance gap between ours and iCaRL~\cite{rebuffi2017icarl}. However, without our weighted-Euclidean regularization, both Norm-iCaRL and iCaRL~\cite{rebuffi2017icarl} exhibit higher variations in phase accuracy than ours (cp. the bottom row of Fig.~\ref{fig:ablation_norm_icarl}).
    
\section{Adding 5 New Classes in Each Phase}
Here we conduct more experiments based on almost the same setting as the one in the main manuscript (i.e. CIFAR-100 and ImageNet datasets, two training scenarios, memory size of 1000 and 2000, as well as the three evaluation metrics), but now the number of new classes in each incremental phase is set to 5 (e.g. in total 20 phases while training from scratch).

\subsection{Incremental Accuracy Comparison}
\label{sec:Evalutation Resluts}
    \begin{figure*}[t]
    \centering
    \subfigure[$M$: 1K, FS]{
    \begin{minipage}[t]{0.235\textwidth}
    \label{fig:5_cifar_ac_a}
    \centering
    \includegraphics[width=1\textwidth]{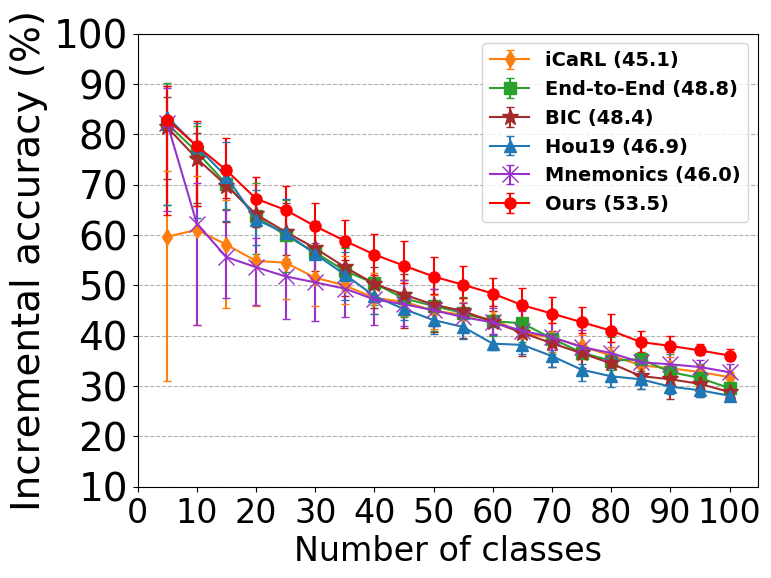}
    \end{minipage}%
    }%
    \subfigure[$M$: 2K, FS]{
    \begin{minipage}[t]{0.235\textwidth}
    \label{fig:5_cifar_ac_b}
    \centering
    \includegraphics[width=1\textwidth]{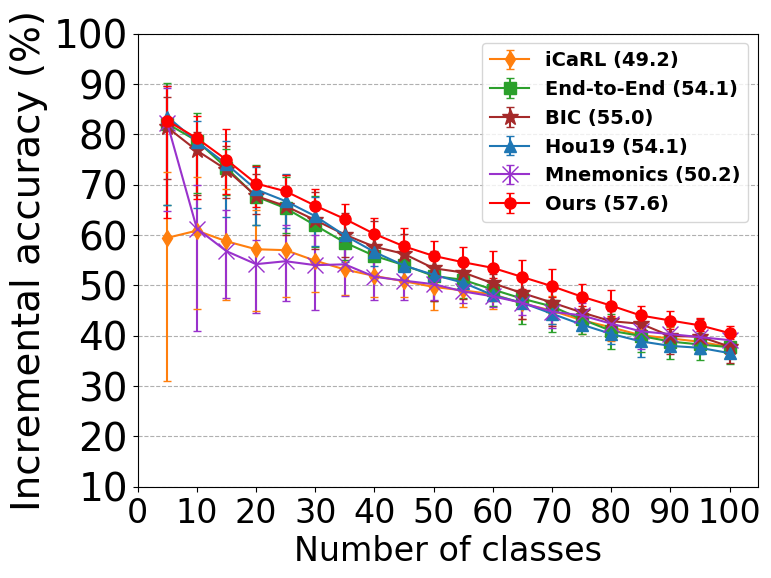}
    \end{minipage}%
    }%
    \subfigure[$M$: 1K, from 50]{
    \begin{minipage}[t]{0.235\textwidth}
    \label{fig:5_cifar_ac_c}
    \centering
    \includegraphics[width=1\textwidth]{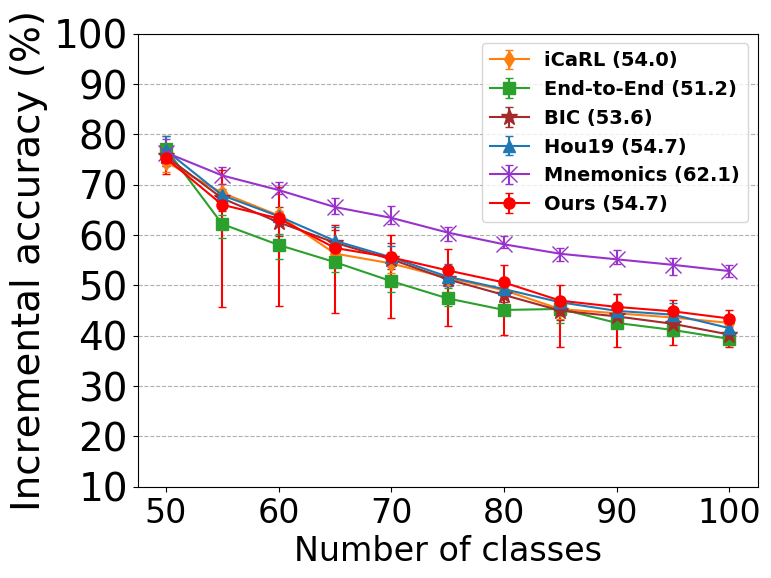}
    \end{minipage}
    }%
    \subfigure[$M$: 2K, from 50]{
    \begin{minipage}[t]{0.235\textwidth}
    \label{fig:5_cifar_ac_d}
    \centering
    \includegraphics[width=1\textwidth]{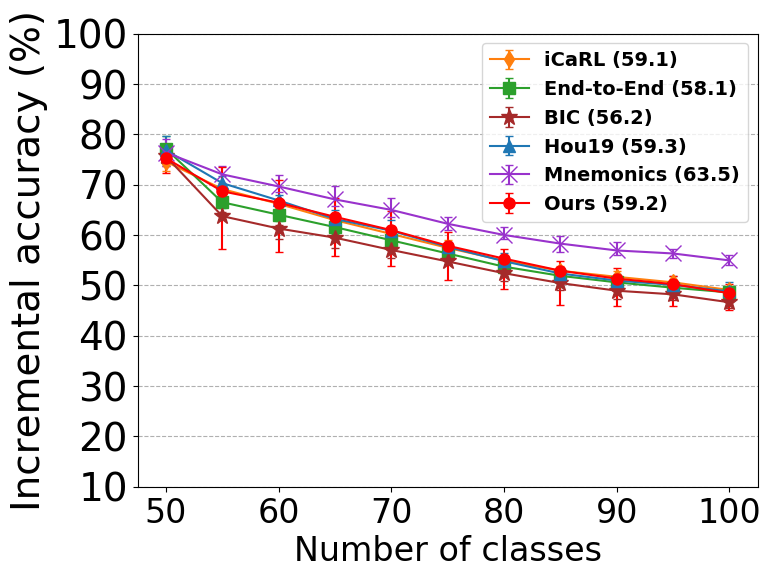}
    \end{minipage}
    }%
    \centering
    
    \subfigure[$M$: 1K, FS]{
    \begin{minipage}[t]{0.235\textwidth}
    \label{fig:5_imagenet_ac_a}
    \centering
    \includegraphics[width=1\textwidth]{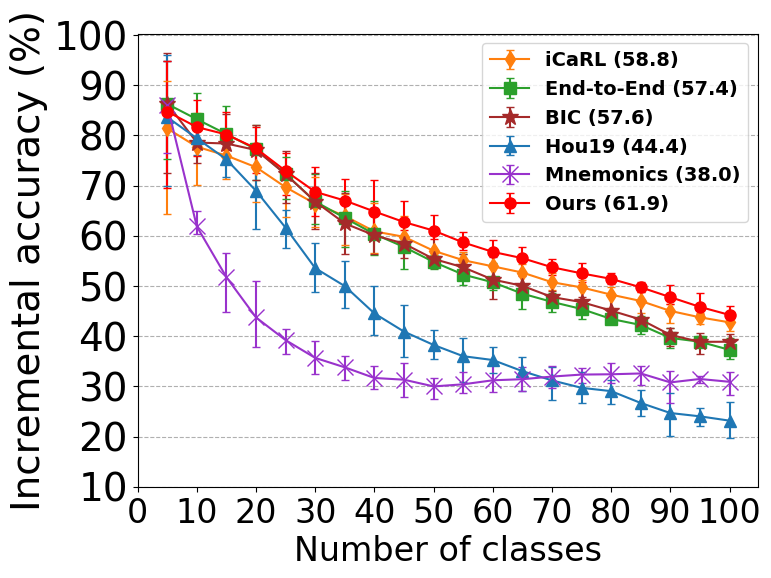}
    \end{minipage}%
    }%
    \subfigure[$M$: 2K, FS]{
    \begin{minipage}[t]{0.235\textwidth}
    \label{fig:5_imagenet_ac_b}
    \centering
    \includegraphics[width=1\textwidth]{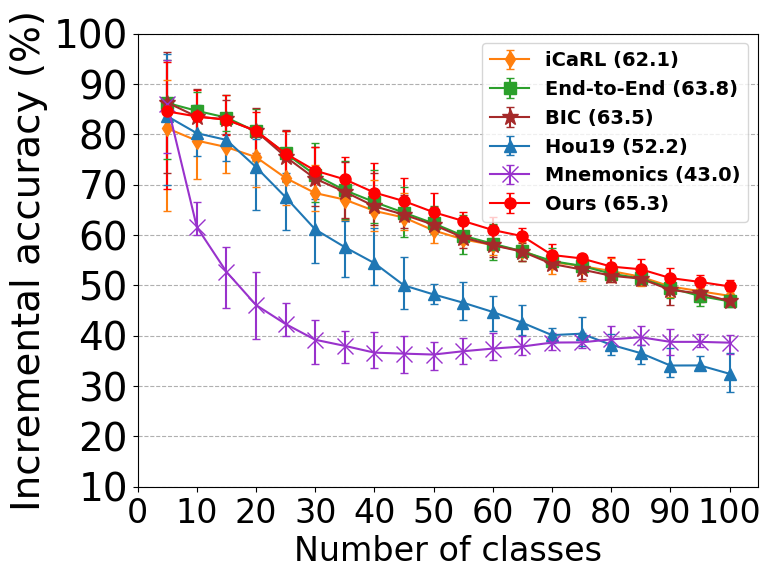}
    \end{minipage}%
    }%
    \subfigure[$M$: 1K, from 50]{
    \begin{minipage}[t]{0.235\textwidth}
    \label{fig:5_imagenet_ac_c}
    \centering
    \includegraphics[width=1\textwidth]{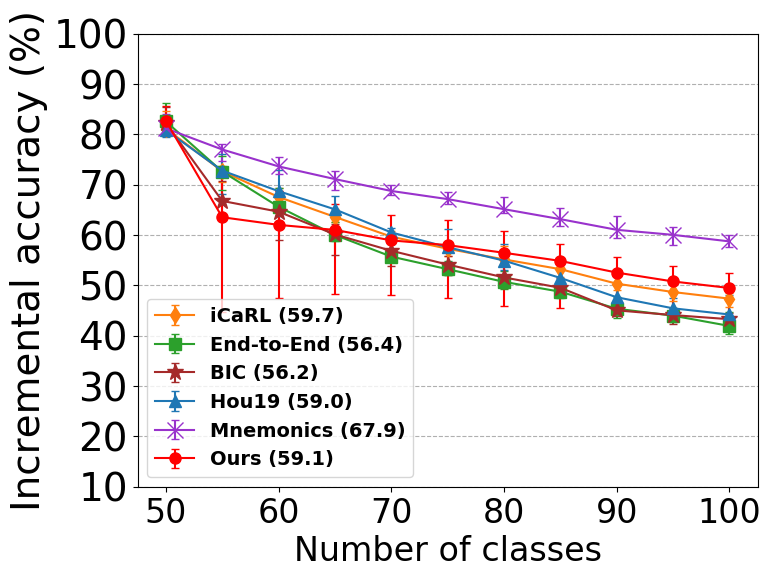}
    \end{minipage}
    }%
    \subfigure[$M$: 2K, from 50]{
    \begin{minipage}[t]{0.235\textwidth}
    \label{fig:5_imagenet_ac_d}
    \centering
    \includegraphics[width=1\textwidth]{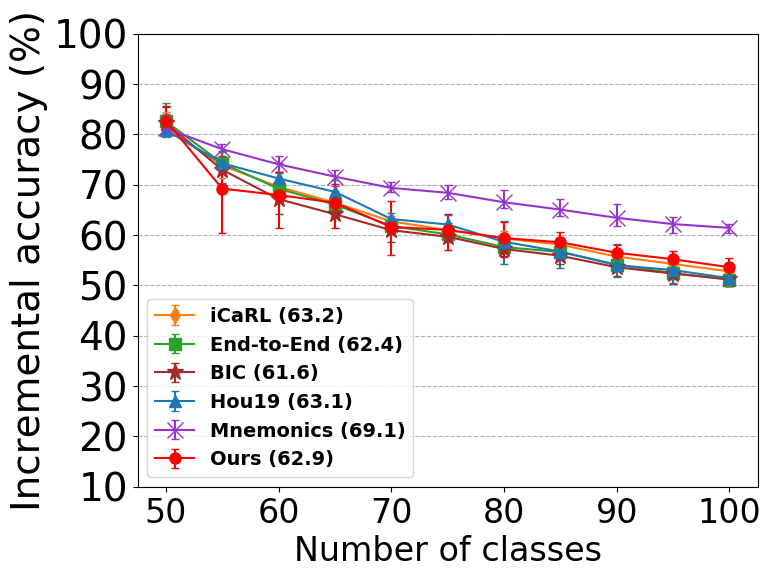}
    \end{minipage}
    }%
    \centering
    \caption{Incremental accuracy on CIFAR-100 (top row) and ImageNet (bottom row), for memory sizes $M=1K,2K$ and training scenarios ``FS" and ``from 50". Average incremental accuracy of each method is shown in parentheses.}
    \label{fig:5_cifar_and_imagenet_ac_a}
    \end{figure*}

    \begin{table}[t]
    \centering
    \renewcommand{\arraystretch}{1}
    \setlength{\tabcolsep}{5.0pt}
    \caption[Comparison of incremental accuracy at the end of training]{Comparison of incremental accuracy at the end of training}
    \label{table:5_accuracy}
    \begin{tabular}{c c c c c c c c c c c c}
    \hline
     Training scenario & \multicolumn{5}{c}{from scratch} && \multicolumn{5}{c}{from 50 classes}\\ \hline
     Dataset & \multicolumn{2}{c}{CIFAR} && \multicolumn{2}{c}{ImageNet}  && \multicolumn{2}{c}{CIFAR} && \multicolumn{2}{c}{ImageNet}\\ \hline
         Memory size & {1K} & {2K} && {1K} & {2K} && {1K} & {2K} && {1K} & {2K}  \\ \hline\hline
    iCaRL~\cite{rebuffi2017icarl}  & 31.8 & 37.7 && 42.7 & 47.9 && 42.5 & 49.4 && 47.4 & 52.8\\ \hline
    End-to-End~\cite{castro2018end}& 29.6 & 37.7 && 37.2  &  46.8 && 39.4 & 48.6 && 41.9 & 51.1 \\ \hline
    BIC~\cite{wu2019large}  & 28.8 & 37.8 && 38.9 &  46.9 && 40.2 & 46.6 && 43.3 & 51.2\\ \hline
    Hou19~\cite{hou2019learning}  & 28.1 & 36.5 &&  23.2&  32.4&&  41.5 &  48.8 && 44.2 & 51.4\\ \hline
    Mnemonics~\cite{liu2020mnemonics} & 32.7 & 39.1 && 30.9 & 38.6 && \textbf{52.8} & \textbf{55.0} && \textbf{58.7} & \textbf{61.4} \\ \hline
    Ours & \textbf{36.1} & \textbf{40.5} && \textbf{44.2} & \textbf{49.8} && 43.4 & 48.4 && 49.5 & 53.6 \\ \hline
    \end{tabular}
    \end{table}
    
    The quantitative evaluation results in terms of incremental accuracy are provided in Fig.~\ref{fig:5_cifar_and_imagenet_ac_a}, while Table~\ref{table:5_accuracy} particularly summarizes the results at the end of the entire incremental-training.

    Similar to the observations we have in the main manuscript, on CIFAR-100 and ImageNet, when learning from scratch, our model outperforms all the baselines with memory size 1000 and 2000. Also, we find again that Hou19~\cite{hou2019learning} and Mnemonics~\cite{liu2020mnemonics} are unable to perform well when the feature extractor is learned from scratch. When starting from 50 classes, ours performs comparably to Mnemonics~\cite{liu2020mnemonics} and outperforms all the other baselines on ImageNet. On CIFAR-100, our model performs comparably to the other baselines with memory size 2000 and achieves the second highest incremental accuracy with memory size 1000.
    When starting from 50 classes, ours outperforms all the other baselines, except Mnemonics~\cite{liu2020mnemonics}, on ImageNet. On CIFAR-100, our model performs comparably to the other baselines with memory size 2000 and achieves the second highest incremental accuracy with memory size 1000.
    
\subsection{Phase Accuracy Comparison}
\label{sec:5_exp_balance_task} 
    \begin{figure}[t]
    \centering
    \subfigure[$M$: 1K, from 50]{
    \begin{minipage}[t]{0.235\textwidth}
    \label{fig:5_task_a}
    \centering
    \includegraphics[width=1\textwidth]{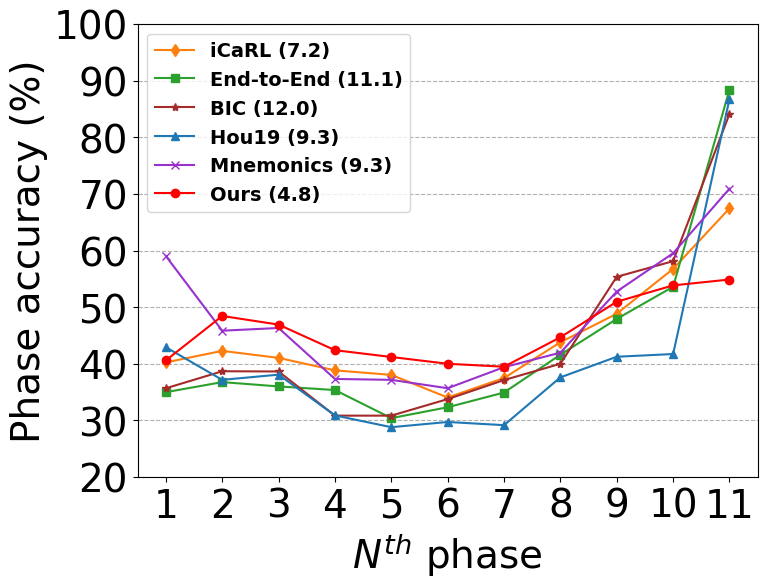}
    \end{minipage}%
    }%
    \subfigure[$M$: 2K, from 50]{
    \begin{minipage}[t]{0.235\textwidth}
    \label{fig:5_task_b}
    \centering
    \includegraphics[width=1\textwidth]{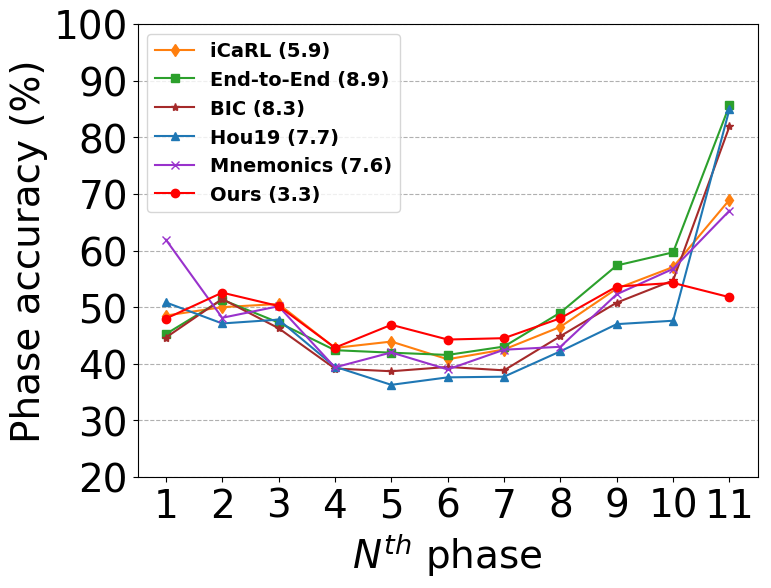}
    \end{minipage}%
    }%
    \subfigure[$M$: 1K, from 50]{
    \begin{minipage}[t]{0.235\textwidth}
    \label{fig:5_task_c}
    \centering
    \includegraphics[width=1\textwidth]{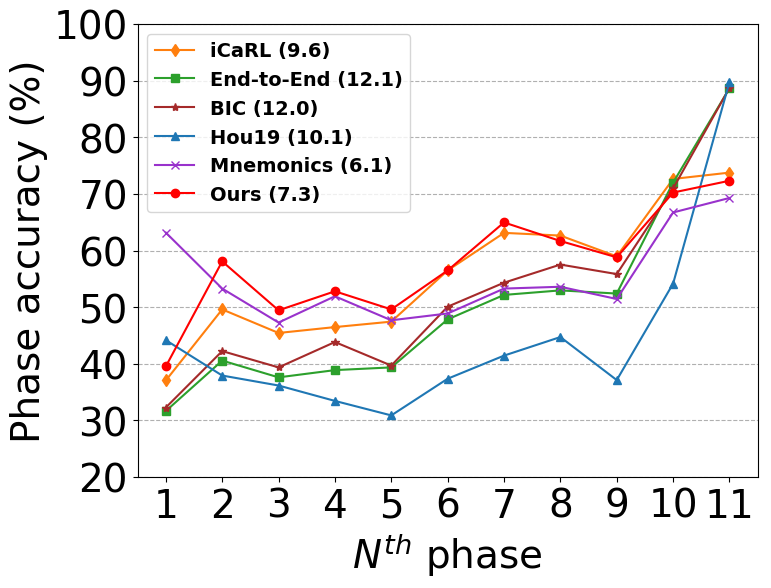}
    \end{minipage}
    }%
    \subfigure[$M$: 2K, from 50]{
    \begin{minipage}[t]{0.235\textwidth}
    \label{fig:5_task_d}
    \centering
    \includegraphics[width=1\textwidth]{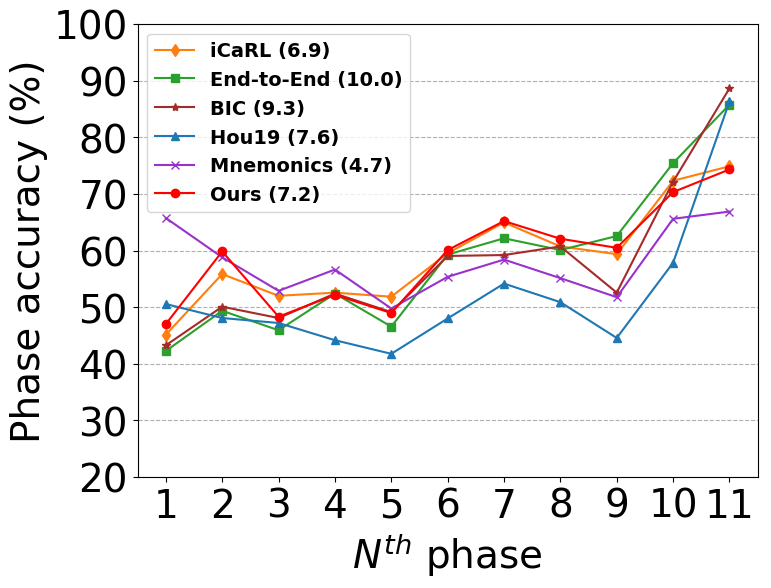}
    \end{minipage}
    }%
    \centering
    \caption{Phase accuracy comparison: (a)(b) are results on CIFAR-100 and (c)(d) on ImageNet. $M$ is the memory size, and the model is pre-trained with 50 classes. The mean absolute deviation in phase accuracy is shown in parentheses.}
    \label{fig:5_balance_accuracy}
    \end{figure}
    
    Fig.~\ref{fig:5_balance_accuracy} presents the phase accuracy for different methods with the training scenario ``from 50", where the baselines perform more closely to our method in terms of incremental accuracy. Shown in parentheses is the mean absolute deviation from the average of each method's phase accuracy. The smaller the deviation, the more balanced the classification accuracy is on different classes. From the figure, our scheme is seen to achieve the minimum mean absolute deviation in phase accuracy on CIFAR-100 and comparable mean absolute deviation to iCaRL~\cite{rebuffi2017icarl} and Mnemonics~\cite{liu2020mnemonics} on ImageNet.

\section{Forgetting Measure}  
    \begin{table}[t]
        \centering
        \renewcommand{\arraystretch}{1}
        \setlength{\tabcolsep}{5.0pt}
        \caption[Comparison of forgetting measure]{Comparison of forgetting measure under the setting that learns 10 new classes in each phase.}
        \label{table:forgetting_measure_10}
        \begin{tabular}{c c c c c c c c c c c c}
        \hline
         Training scenario & \multicolumn{5}{c}{from scratch} && \multicolumn{5}{c}{from 50 classes}\\ \hline
         Dataset & \multicolumn{2}{c}{CIFAR} && \multicolumn{2}{c}{ImageNet}  && \multicolumn{2}{c}{CIFAR} && \multicolumn{2}{c}{ImageNet}\\ \hline
             Memory size & {1K} & {2K} && {1K} & {2K} && {1K} & {2K} && {1K} & {2K}  \\ \hline\hline
        iCaRL~\cite{rebuffi2017icarl}  & \textbf{24.6} & 20.6 && \textbf{22.3} & \textbf{20.6} && 24.6 & 20.4 && 24.1 & 20.4\\ \hline
        End-to-End~\cite{castro2018end}& 41.6 & 32.9 && 39.1  & 31.0 && 38.2 & 29.1 && 37.2 & 28.4 \\ \hline
        BIC~\cite{wu2019large}  & 37.3 & 28.6 && 36.1 & 31.4  && 33.5 & 26.4 && 33.9 & 27.3\\ \hline
        Hou19~\cite{hou2019learning}  & 27.3 & \textbf{16.8} && 38.7 & 28.8 && \textbf{17.2} & \textbf{11.1} && 20.4 & \textbf{13.4}\\ \hline
        Mnemonics~\cite{liu2020mnemonics} & 39.9 & 33.1 && 38.4 & 31.7 && 22.8 & 18.9 && \textbf{15.9} & 14.0 \\ \hline
        Ours & 25.0 & 22.3 && 24.6 & 23.6 && 17.7 & 16.1 && 18.4 & 17.8 \\ \hline
    
    \end{tabular}
    \end{table}
    
    \begin{table}[t]
        \centering
        \renewcommand{\arraystretch}{1}
        \setlength{\tabcolsep}{5.0pt}
        \caption[Comparison of forgetting measure]{Comparison of forgetting measure under the setting that learns 5 new classes in each phase}
        \label{table:forgetting_measure_5}
        \begin{tabular}{c c c c c c c c c c c c}
        \hline
         Training scenario & \multicolumn{5}{c}{from scratch} && \multicolumn{5}{c}{from 50 classes}\\ \hline
         Dataset & \multicolumn{2}{c}{CIFAR} && \multicolumn{2}{c}{ImageNet}  && \multicolumn{2}{c}{CIFAR} && \multicolumn{2}{c}{ImageNet}\\ \hline
             Memory size & {1K} & {2K} && {1K} & {2K} && {1K} & {2K} && {1K} & {2K}  \\ \hline\hline
        iCaRL~\cite{rebuffi2017icarl}  & 34.5 & 29.4 && \textbf{34.1} & \textbf{30.6} && 32.7 & 26.7 && 36.3 & 30.5\\ \hline
        End-to-End~\cite{castro2018end}& 55.6 & 44.2 &&  56.1 & 43.8 && 47.4 & 36.2 && 49.4 & 37.9 \\ \hline
        BIC~\cite{wu2019large}  & 50.4 & 40.9 && 54.6 &  45.7 && 43.2 & 35.9 && 48.5 & 39.9\\ \hline
        Hou19~\cite{hou2019learning}  & 61.7 & 51.6 && 72.5 & 63.0 &&  46.6 &  38.2&& 50.3 & 41.8\\ \hline
        Mnemonics~\cite{liu2020mnemonics} & 47.4 & 39.4 && 49.8 & 39.5 && \textbf{24.8} & \textbf{20.0} && \textbf{20.2} & \textbf{20.3} \\ \hline
        Ours & \textbf{28.0} & \textbf{23.9} && 35.6 & \textbf{30.6} && 26.4 & 21.8 && 32.1 & 27.8 \\ \hline
        \end{tabular}
    \end{table}
    In this section, we present the results of forgetting measure~\cite{chaudhry2018riemannian} on every class at the end of training, in order to understand to what degree the competing methods forget the previously learned classes. Generally, the lower the forgetting measure, the less the catastrophic forgetting. Table~\ref{table:forgetting_measure_10} summarizes the results for the setting that learns 10 new classes in each training phase. It shows that our method has low forgetting measure and performs comparably to iCaRL~\cite{rebuffi2017icarl}, Hou19~\cite{hou2019learning} and Mnemonics~\cite{liu2020mnemonics}.
    
    Table~\ref{table:forgetting_measure_5} further provides the results for the setting of learning 5 new classes in each phase. In this setting, the model needs to be updated more frequently and tends to forget more easily what it has learned. On CIFAR-100, ours achieves the lowest forgetting measure, when learning from scratch. Moreover, it has the second lowest forgetting measure with the training scenario ``from 50". On ImageNet, iCaRL~\cite{rebuffi2017icarl} performs comparably to ours, when learning from scratch. Since Hou19~\cite{hou2019learning} and Mnemonics~\cite{liu2020mnemonics} are unable to perform well when learning from scratch, they suffer from serious catastrophic forgetting, showing large forgetting measure. When learning from 50 classes, ours performs comparably to Mnemonics~\cite{liu2020mnemonics} and outperforms the other baselines.
    
    To sum up, our method has the ability to preserve well the knowledge of old classes under various settings, while showing high incremental accuracy in learning new classes and more balanced phase accuracy.

%% file: 0038.bbl
\begin{thebibliography}{10}

\bibitem{rebuffi2017icarl}
Rebuffi, S.A., Kolesnikov, A., Sperl, G., Lampert, C.H.:
\newblock icarl: Incremental classifier and representation learning.
\newblock In: IEEE Conference on Computer Vision and Pattern Recognition
  (CVPR). (2017)

\bibitem{aljundi2018memory}
Aljundi, R., Babiloni, F., Elhoseiny, M., Rohrbach, M., Tuytelaars, T.:
\newblock Memory aware synapses: Learning what (not) to forget.
\newblock In: European Conference on Computer Vision (ECCV). (2018)

\bibitem{kirkpatrick2017overcoming}
Kirkpatrick, J., Pascanu, R., Rabinowitz, N., Veness, J., Desjardins, G., Rusu,
  A.A., Milan, K., Quan, J., Ramalho, T., Grabska-Barwinska, A.,  et~al.:
\newblock Overcoming catastrophic forgetting in neural networks.
\newblock Proceedings of the National Academy of Sciences (PNAS) (2017)

\bibitem{li2017learning}
Li, Z., Hoiem, D.:
\newblock Learning without forgetting.
\newblock IEEE Transactions on Pattern Analysis and Machine Intelligence
  (TPAMI) (2017)

\bibitem{hou2018lifelong}
Hou, S., Pan, X., Change~Loy, C., Wang, Z., Lin, D.:
\newblock Lifelong learning via progressive distillation and retrospection.
\newblock In: European Conference on Computer Vision (ECCV). (2018)

\bibitem{aljundi2017expert}
Aljundi, R., Chakravarty, P., Tuytelaars, T.:
\newblock Expert gate: Lifelong learning with a network of experts.
\newblock In: IEEE Conference on Computer Vision and Pattern Recognition
  (CVPR). (2017)

\bibitem{lopez2017gradient}
Lopez-Paz, D., Ranzato, M.:
\newblock Gradient episodic memory for continual learning.
\newblock In: Advances in Neural Information Processing Systems (NIPS). (2017)

\bibitem{chaudhry2018efficient}
Chaudhry, A., Ranzato, M., Rohrbach, M., Elhoseiny, M.:
\newblock Efficient lifelong learning with a-gem.
\newblock In: International Conference on Learning Representations (ICLR).
  (2019)

\bibitem{mccloskey1989catastrophic}
McCloskey, M., Cohen, N.J.:
\newblock Catastrophic interference in connectionist networks: The sequential
  learning problem.
\newblock In: Psychology of Learning and Motivation.
\newblock (1989)

\bibitem{zenke2017continual}
Zenke, F., Poole, B., Ganguli, S.:
\newblock Continual learning through synaptic intelligence.
\newblock In: International Conference on Machine Learning (ICML). (2017)

\bibitem{chaudhry2018riemannian}
Chaudhry, A., Dokania, P.K., Ajanthan, T., Torr, P.H.:
\newblock Riemannian walk for incremental learning: Understanding forgetting
  and intransigence.
\newblock In: European Conference on Computer Vision (ECCV). (2018)

\bibitem{aljundi2018selfless}
Aljundi, R., Rohrbach, M., Tuytelaars, T.:
\newblock Selfless sequential learning.
\newblock In: International Conference on Learning Representations (ICLR).
  (2019)

\bibitem{mallya2018piggyback}
Mallya, A., Davis, D., Lazebnik, S.:
\newblock Piggyback: Adapting a single network to multiple tasks by learning to
  mask weights.
\newblock In: European Conference on Computer Vision (ECCV). (2018)

\bibitem{mallya2018packnet}
Mallya, A., Lazebnik, S.:
\newblock Packnet: Adding multiple tasks to a single network by iterative
  pruning.
\newblock In: IEEE Conference on Computer Vision and Pattern Recognition
  (CVPR). (2018)

\bibitem{hinton2015distilling}
Hinton, G., Vinyals, O., Dean, J.:
\newblock Distilling the knowledge in a neural network.
\newblock In: NIPS Workshop on Deep Learning and Representation Learning.
  (2014)

\bibitem{belouadah2018deesil}
Belouadah, E., Popescu, A.:
\newblock Deesil: Deep-shallow incremental learning.
\newblock In: European Conference on Computer Vision (ECCV). (2018)

\bibitem{castro2018end}
Castro, F.M., Mar{\'\i}n-Jim{\'e}nez, M.J., Guil, N., Schmid, C., Alahari, K.:
\newblock End-to-end incremental learning.
\newblock In: European Conference on Computer Vision (ECCV). (2018)

\bibitem{hou2019learning}
Hou, S., Pan, X., Loy, C.C., Wang, Z., Lin, D.:
\newblock Learning a unified classifier incrementally via rebalancing.
\newblock In: IEEE Conference on Computer Vision and Pattern Recognition
  (CVPR). (2019)

\bibitem{jung2018less}
Jung, H., Ju, J., Jung, M., Kim, J.:
\newblock Less-forgetful learning for domain expansion in deep neural networks.
\newblock In: AAAI Conference on Artificial Intelligence (AAAI). (2018)

\bibitem{wu2019large}
Wu, Y., Chen, Y., Wang, L., Ye, Y., Liu, Z., Guo, Y., Fu, Y.:
\newblock Large scale incremental learning.
\newblock In: IEEE Conference on Computer Vision and Pattern Recognition
  (CVPR). (2019)

\bibitem{liu2020mnemonics}
Liu, Y., Su, Y., Liu, A.A., Schiele, B., Sun, Q.:
\newblock Mnemonics training: Multi-class incremental learning without
  forgetting.
\newblock In: IEEE Conference on Computer Vision and Pattern Recognition
  (CVPR). (2020)

\bibitem{goodfellow2014generative}
Goodfellow, I., Pouget-Abadie, J., Mirza, M., Xu, B., Warde-Farley, D., Ozair,
  S., Courville, A., Bengio, Y.:
\newblock Generative adversarial nets.
\newblock In: Advances in Neural Information Processing Systems (NIPS). (2014)

\bibitem{shin2017continual}
Shin, H., Lee, J.K., Kim, J., Kim, J.:
\newblock Continual learning with deep generative replay.
\newblock In: Advances in Neural Information Processing Systems (NIPS). (2017)

\bibitem{he2018exemplar}
He, C., Wang, R., Shan, S., Chen, X.:
\newblock Exemplar-supported generative reproduction for class incremental
  learning.
\newblock In: British Machine Vision Conference (BMVC). (2018)

\bibitem{ostapenko2019learning}
Ostapenko, O., Puscas, M., Klein, T., Jahnichen, P., Nabi, M.:
\newblock Learning to remember: A synaptic plasticity driven framework for
  continual learning.
\newblock In: IEEE Conference on Computer Vision and Pattern Recognition
  (CVPR). (2019)

\bibitem{lecun1998gradient}
LeCun, Y., Bottou, L., Bengio, Y., Haffner, P.:
\newblock Gradient-based learning applied to document recognition.
\newblock Proceedings of the IEEE (1998)

\bibitem{Taitelbaum2018AddingNC}
Taitelbaum, H., Ben-Reuven, E., Goldberger, J.:
\newblock Adding new classes without access to the original training data with
  applications to language identification.
\newblock In: Annual Conference of the International Speech Communication
  Association (INTERSPEECH). (2018)

\bibitem{krizhevsky2009learning}
Krizhevsky, A.:
\newblock Learning multiple layers of features from tiny images.
\newblock Technical report (2009)

\bibitem{deng2009imagenet}
Deng, J., Dong, W., Socher, R., Li, L.J., Li, K., Fei-Fei, L.:
\newblock Imagenet: A large-scale hierarchical image database.
\newblock In: IEEE Conference on Computer Vision and Pattern Recognition
  (CVPR). (2009)

\bibitem{he2016deep}
He, K., Zhang, X., Ren, S., Sun, J.:
\newblock Deep residual learning for image recognition.
\newblock In: IEEE Conference on Computer Vision and Pattern Recognition
  (CVPR). (2016)

\bibitem{welling2009herding}
Welling, M.:
\newblock Herding dynamical weights to learn.
\newblock In: International Conference on Machine Learning (ICML). (2009)

\end{thebibliography}
